%% file: main.tex
\documentclass{article}

\usepackage[dvipsnames]{xcolor}
\usepackage{caption, subcaption}
\usepackage{microtype}
\usepackage{graphicx}
\usepackage{booktabs} %
\usepackage{amsmath, amsfonts, todonotes, bm, cases}

\usepackage{tikz}
\usetikzlibrary{bayesnet, arrows}
\usetikzlibrary{shapes.geometric, backgrounds, shapes.misc, fadings}
\usepackage{pgfplots} %
\usepackage{array} %
\usepackage{multirow}
\usepackage{siunitx}
\usepackage{graphbox} %
\usepackage{algorithm}
\usepackage{algorithmicx}
\usepackage{algpseudocode}

\usepackage{enumitem}
\setlist[itemize,1]{leftmargin=\dimexpr 26pt-.1in}

\usepackage{url}

\usepackage{breakurl}
\usepackage[breaklinks]{hyperref}

\makeatletter
\newcommand\nocaption{%
    \renewcommand\p@subfigure{}
    \renewcommand\thesubfigure{\thefigure\alph{subfigure}}
}
\makeatother

\definecolor{mycolour1}{RGB}{0, 119, 187}
\definecolor{mycolour2}{RGB}{0, 153, 136}
\definecolor{mycolour3}{RGB}{204, 51, 17}
\definecolor{mycolour4}{RGB}{238, 119, 51}

\def\pthx_z{p_{\boldsymbol \theta}({\bf x}|{\bf z})}

\usepackage[nonatbib,final]{neurips_2020}
\usepackage[numbers]{natbib}
\pgfplotsset{compat=1.16}

\usepackage[utf8]{inputenc} %
\usepackage[T1]{fontenc}    %
\usepackage{nicefrac}       %

\title{Hierarchical Quantized Autoencoders}

\author{%
  Will Williams\thanks{Equal contribution.} \\
  \texttt{willw@speechmatics.com} \\
  \And
  Sam Ringer$^*$ \\
  \texttt{samr@speechmatics.com} \\
  \And
  John Hughes \\
  \texttt{johnh@speechmatics.com} \\
  \And
  Tom Ash \\
  \texttt{toma@speechmatics.com} \\
  \And
  David MacLeod \\
  \texttt{davidma@speechmatics.com} \\
  \And
  Jamie Dougherty \\
  \texttt{jamied@speechmatics.com} \\
}

\begin{document}

\maketitle

\vspace{-0.5em}

\begin{abstract}
Despite progress in training neural networks for lossy image compression, current approaches fail to maintain both perceptual quality and abstract features at very low bitrates. Encouraged by recent success in learning discrete representations with Vector Quantized Variational Autoencoders (VQ-VAEs), we motivate the use of a hierarchy of VQ-VAEs to attain high factors of compression. We show that the combination of stochastic quantization and hierarchical latent structure aids likelihood-based image compression. This leads us to introduce a novel objective for training hierarchical VQ-VAEs. Our resulting scheme produces a Markovian series of latent variables that reconstruct images of high-perceptual quality which retain semantically meaningful features. We provide qualitative and quantitative evaluations on the CelebA and MNIST datasets.

\end{abstract}

\section{Introduction}

\label{intro}

The internet age relies on lossy compression algorithms that transmit information at low bitrates. These algorithms are typically analysed through the rate-distortion trade-off, originally posited by \citet{shannon1959coding}. When performing lossy compression at extremely low bit rates, obtaining low distortions often results in reconstructions of very low perceptual quality \citep{blau2018perception,blau2019rethinking,tschannen2018deep}. For modern lossy compression, high perceptual quality of reconstructions is often more desirable than low distortions. This work investigates good performance on this rate-perception tradeoff as opposed to more standard rate-distortion trade offs, with a focus on the low-rate regime.

At low bitrates it is desirable to communicate only high-level concepts and offload the `filling in' of details to a powerful decoder \cite{tschannen2018deep}. Neural Networks present a promising avenue since they are flexible enough to learn the complex transformations required to both capture such high-level concepts and reconstruct in a convincing way that avoids artifacts \cite{generativeCompression,gregor2016towards,johnston2019computationally}.

Variational Autoencoders (VAEs \cite{vae}) are latent variable Neural Network models that have made significant strides in lossy image compression \cite{CompAE, soft2hard}. However, due to a combination of a poor likelihood function and a sub-optimal variational posterior \cite{tamingvaes, PGAE}, reconstructions can look blurred and unrealistic \cite{towardsvae, betaVAE}. There have been many attempts to construct hierarchical forms of both VAEs and Vector Quantized Variational Autoencoders (VQ-VAEs), however perceptual quality is frequently sacrificed at low-rates, and has only recently been made viable with methods that require large autoregressive decoders \cite{HAMS, VQVAE2}. Solutions to this problem then take two forms: either augmenting the likelihood model, for instance, by using adversarial methods \cite{tschannen2018deep} or improving the structure of the posterior/latent space \cite{PGAE, BrokenElbo}. However, at low rates both solutions struggle to match the realism of implicit generative models \cite{goodfellow2014generative}. 

\begin{table*}[htbp]
\caption{CelebA interpolations of the HQA encoder output $z_e$ in the 9 bit 8x8 latent space. The original 64x64 images are shown on the left and right. The center images are the resulting decodes when using 8 linearly interpolated points between the $z_e$ of the original images. Compression is from 98,304 to 576 bits (171x compression).}

\tabcolsep=0.11cm
\small
\centering
\begin{tabular}{c|c c c c c c c c|c}
\includegraphics[height=1.4cm,keepaspectratio]{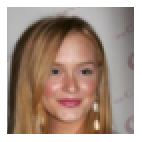} 	&   \multicolumn{8}{c|}{\includegraphics[height=1.4cm,keepaspectratio]{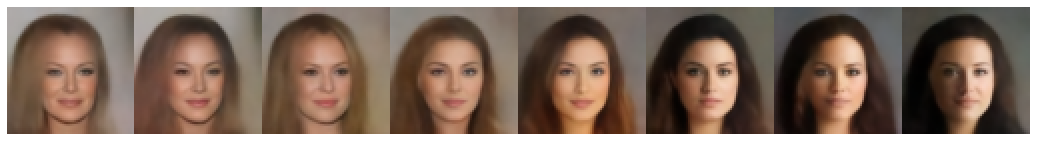}}  &  \includegraphics[height=1.4cm,keepaspectratio]{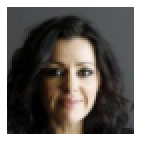} 	\\
	\includegraphics[height=1.45cm,keepaspectratio]{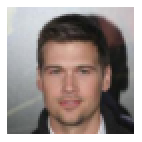} 	&   \multicolumn{8}{c|}{\includegraphics[height=1.4cm,keepaspectratio]{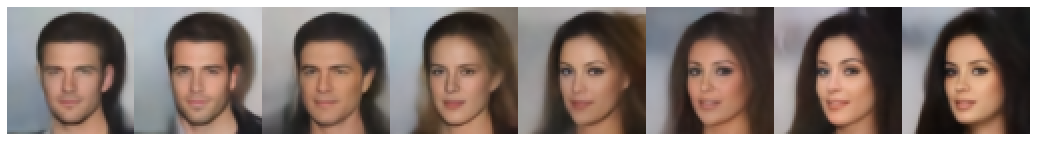}}  &  \includegraphics[height=1.4cm,keepaspectratio]{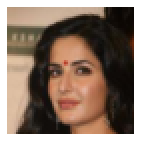} \\ 
\includegraphics[height=1.4cm,keepaspectratio]{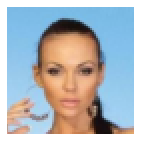} 	&   \multicolumn{8}{c|}{\includegraphics[height=1.4cm,keepaspectratio]{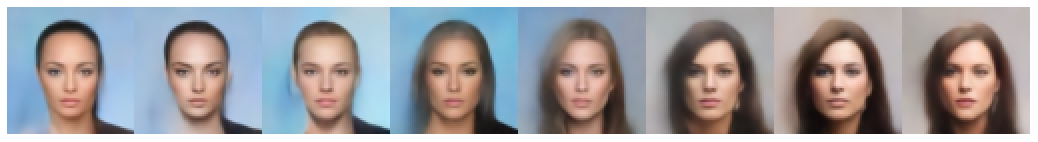}}  &  \includegraphics[height=1.4cm,keepaspectratio]{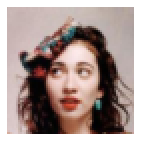} \\ 
\includegraphics[height=1.4cm,keepaspectratio]{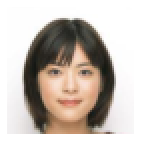} 	&   \multicolumn{8}{c|}{\includegraphics[height=1.4cm,keepaspectratio]{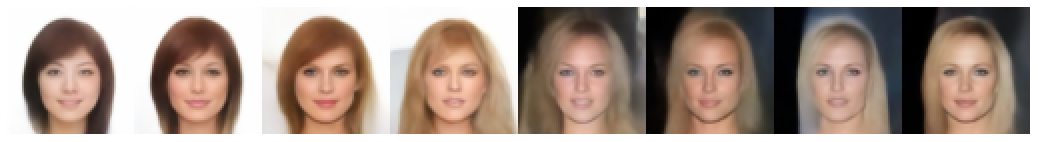}}  &  \includegraphics[height=1.4cm,keepaspectratio]{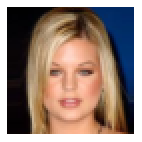} \\ 
\end{tabular}

\label{tab:interp-celeba}
\end{table*}

To address these issues, we build from previous work on heirarchical VQ-VAEs and introduce\footnote{Code available at \url{https://github.com/speechmatics/hqa}} the `Hierarchical Quantized Autoencoder' (HQA). Our system implicitly gives rise to many of the qualities of explicit perceptual losses and furnishes the practitioner with a repeatable operation of learned-compression that can be trained greedily.

Our key contributions are as follows:
\begin{itemize}
    \itemsep0em
    \item We introduce new analysis as to why probabilistic quantized hierarchies are particularly well-suited to optimising the perception-rate tradeoff when performing extreme lossy compression.
    
    \item We propose a new scheme (HQA) for extreme lossy compression. HQA  exploits probabilistic forms of VQ-VAE's commitment and codebook losses and uses a novel objective for training hierarchical VQ-VAEs. This objective leads to higher layers implicitly reconstructing the full posterior of the layer below, as opposed to samples from this posterior.
    
    \item We show that HQA can produce reconstructions of high perceptual quality at very low rates using only simple feedforward decoders, where as related methods require autoregressive decoders.
\end{itemize}

\section{Related Work}
\subsection{Lossy Compression and the Rate-Perception Trade-off}
Shannon's rate-distortion theory of lossy compression makes no claims about perceptual quality. \citet{blau2019rethinking} show that optimising for distortion necessitates a trade off with perceptual quality, particularly at extremely low rates. This move to focus on perceptual quality has motivated the introduction of perceptual losses \cite{toderici2017full,balle2018variational,generativeCompression,patel2019deep} which are heuristically defined and attempt to capture different aspects of human-perceived perceptual quality. Our work naturally gives rise to losses at different levels of abstraction which have a similar effect as perceptual losses but which are less heuristically defined and encourage abstract semantic categories to be captured. This leads to good performance on the rate-perception task on which we focus.

\citet{blau2019rethinking} extend lossy compression to allow for stochastic decodes. Prior work \cite{tschannen2018deep, agustsson2019generative} notes that to achieve good perceptual quality at extreme rates, stochastic decoders are essential. Stochasticity has previously been introduced in an ad-hoc manner by injecting a noise vector into the decoder alongside the code. This is the same strategy used by most conditional generative models. However, this artificial introduction of stochasticity is problematic as the decoder often learns to ignore the noise vector completely \cite{zhu2017toward,isola2017image}. HQA parameterizes distributions over codes at different layers of abstraction, each of which can be sampled from in turn. This introduces stochasticity in a more natural and nonrestrictive manner.

\subsection{VAE hierarchies}
\label{sec:vae_hierarchy_relatedwork}
Our work is most closely related to \citet{gregor2016towards}, where a VAE-based hierarchy is constructed in an attempt to capture increasingly abstract concepts. Similarly, we only need to transmit top-level latents of a hierarchical model for use as a lossy code. However, their scheme relies on expensive iterative computation to decode latents and they struggle empirically to maintain perceptual quality at low rates. They rely on iterative refinement to obtain sharpness whereas our scheme can obtain a sharp and credible reconstruction with a single computational pass through the network. Additionally, they can only transmit a subset of the higher levels in the hierarchy, whereas each layer in our hierarchy represents a fully independent lossy code which can be transmitted at a fixed rate.

VQ-VAE-2 \cite{VQVAE2} introduces a hierarchy of VQ-VAEs and is trained using a two stage procedure. During the first stage all VQ-VAEs are trained jointly under one objective. During the second stage, large autoregressive decoders are trained and replace the original decoders. Although introduced as a generative model, the system after each of these stages can potentially be used for lossy compression. After the first stage, the structure of VQ-VAE-2 is such that the latents from \emph{all} layers are required for image reconstruction. Therefore, all latents must be transmitted to perform lossy compression, making low-rate compression near impossible. The system after the second stage of training is more suitable for lossy compression as only the highest level latents need transmitting. However, the new decoders then dominate the parameter count in the final model by several orders of magnitude and their autoregressive nature lead to computationally burdensome reconstruction times. Additionally, for each fixed compression rate, a whole new VQ-VAE-2 must be trained through both stages. Instead, we look to compare against schemes that use simple feedforward decoders and that have feasible scaling properties across many bitrates.

One such scheme is the Hierarchical Autoregressive Model \cite{HAMS} (denoted HAMs). Similar to VQ-VAE-2, HAMs train a hierarchy of VQ-VAEs in a two step procedure, with the second step training a series of autoregressive auxillary decoders. In contrast to VQ-VAE-2, the hierarchy obtained after the first stage is suitable for extreme lossy compression as only the top level latents need to be transmitted and only simple feedforward decoders are used. In contrast to HQA, each layer of HAMs produces a deterministic posterior and each decoder is trained with a cross-entropy loss over the code indices of the layer below.

\section{Background}
\subsection{VQ-VAE}

\label{background}
VQ-VAEs~\cite{VQVAE,VQVAE2} model high dimensional data $x$ with low-dimensional discrete latents $z$. A likelihood function $p_\theta(x|z)$ is parameterized with a decoder that maps from latent space to observation space. A uniform prior distribution $p(z)$ is defined over a discrete space of latent codes. As in the variational inference framework~\cite{neurVI}, an approximate posterior is defined over the latents:
\begin{align}
\small
    q_{\phi}(z = k|x) = \begin{cases}1 \quad\text{for} \enspace k = \text{argmin}_j ||z_e(x) - e_j||_2\\ 0\quad \text{otherwise}\end{cases}
\end{align}
The codebook $(e_i)_{i=1}^N$ enumerates a list of vectors and an encoder $z_e(x)$ maps into latent space. A vector quantization operation then maps the encoded observation to the nearest code. During training the encoder and decoder are trained jointly to minimize the loss:
\begin{equation}\label{eqn:vqvaeloss}
\small
  - \log p_{\theta}(x|z=k) + ||\text{\textit{sg}}[z_e(x)] - e_k||_{2}^2 \\+ \beta||z_e(x) - \text{\textit{sg}}[e_k]||_2^2 \: ,
\end{equation}
where \textit{sg} is a stop gradient operator. The first term is referred to as the \textit{reconstruction loss}, the second term is the \textit{codebook loss} and the final term is the \textit{commitment loss}. In practice, the codes $e_k$ are learnt via an online exponential moving average version of k-means.

\section{Lossy Compression Using Quantized Hierarchies}
\label{motivation}

Lossy compression schemes will invariably use some form of quantization to select codes for transmission. This section examines the behaviour of quantization-based models trained using maximum-likelihood.

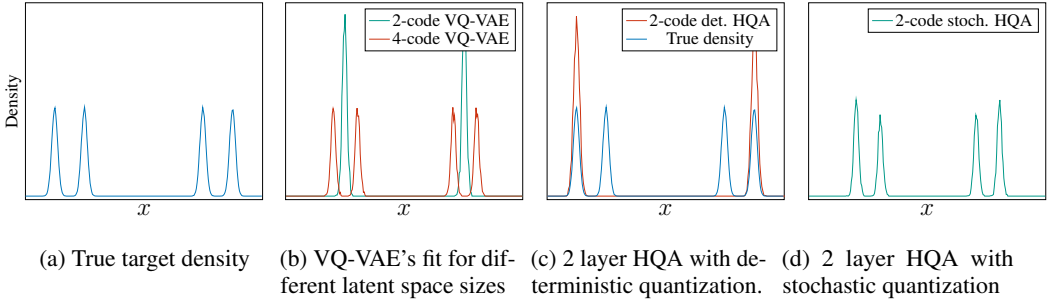
\begin{figure*}[!h]
    \centering
    \begin{subfigure}[t]{.25\textwidth}
        \centering
        \input{tikzdiagrams/toy_data_graph_1.tex}
    \end{subfigure}
    ~
    \begin{subfigure}[t]{.23\textwidth}
        \centering
        \input{tikzdiagrams/toy_data_graph_2.tex}
    \end{subfigure}
    ~
    \begin{subfigure}[t]{.23\textwidth}
        \centering
        \input{tikzdiagrams/toy_data_graph_4.tex}
    \end{subfigure}
    ~
    \begin{subfigure}[t]{.23\textwidth}
        \centering
        \input{tikzdiagrams/toy_data_graph_3.tex}
    \end{subfigure}
    \newline
    \begin{subfigure}[t]{.22\textwidth}
        \caption{True target density}
         \label{fig:toy_distribution}
    \end{subfigure}
    ~
    \begin{subfigure}[t]{.22\textwidth}
        \caption{VQ-VAE's fit for different latent space sizes}
         \label{fig:toy_single_layer_vqvae}
    \end{subfigure}
    ~
    \begin{subfigure}[t]{.22\textwidth}
        \caption{2 layer HQA with deterministic quantization.}
         \label{fig:toy_deterministic}
    \end{subfigure}
    ~
    \begin{subfigure}[t]{.22\textwidth}
        \caption{2 layer HQA with stochastic quantization}
        \label{fig:toy_stochastic}
    \end{subfigure}
    \vspace{-1em}
    \caption{Modelling a simple multi-modal distribution using different forms of hierarchies. The HQA system uses the pre-trained 4-code VQ-VAE from Figure \ref{fig:toy_single_layer_vqvae} and adds a 2-code VQ-VAE on top. Note, for HQA, only the top-layer codes count for transmission since the lower level codes are generated during decoding.}
    \vspace{-1em}
\end{figure*}

\subsection{Illustrative task}
Consider performing standard lossy compression on datapoints sampled from the distribution shown in Figure~\ref{fig:toy_distribution}. Each datapoint is encoded to an encoding consisting of only a small number of bits. Each encoding is then decoded to obtain an imperfect (lossy) reconstruction of the original datapoint. We desire a lossy compression system that shows the following behaviour:

\textbf{Low Bitrates} The encoding of each datapoint should consist of as few bits as possible.

\textbf{Realism} The reconstruction of each datapoint should \textit{not} take on a value that has low probability under the original distribution. For the distribution in Figure~\ref{fig:toy_distribution}, this corresponds to regions outside of the four modes. We term such reconstructions \textit{unrealistic}. In other words, it should never be the case that a reconstruction is \textit{clearly} not from the original distribution. A link can be drawn between these areas of low probability in the original data distribution and the blurry/unrealistic samples often seen when using VAEs for reconstruction tasks.

\subsection{Single Layer VQ-VAE}
\subsubsection{4-Code VQ-VAE}

We begin by using a VQ-VAE to compress and reconstruct samples from the density shown in Figure~\ref{fig:toy_distribution}. We first train a VQ-VAE that uses a latent space of 4 codewords. The encodings produced by this VQ-VAE will therefore each be of size 2 bits ($= \text{log}_24$). The red trace in Figure~\ref{fig:toy_single_layer_vqvae} shows the density of the reconstructions from this 4-code VQ-VAE. It is a perfect match of the density function that the original datapoints were sampled from. There are no \textit{unrealistic} reconstructions as all reconstructed datapoints fall in regions of high density under the original distribution.

\subsubsection{2-Code VQ-VAE}

We now fit a VQ-VAE with a 2 codeword (1 bit) latent space to the original density. The green trace of Figure~\ref{fig:toy_single_layer_vqvae} shows the result. The mode-covering behaviour shown by this VQ-VAE causes reconstructions to fall in regions of low probability under the original distribution. Therefore, nearly all reconstructions are \textit{unrealistic}. This mode-covering is a well known pathology of all likelihood-based models trained using the asymmetric divergence $\text{KL}[p_{\theta}(x)||p(x)]$. \cite{Adaptive_Density_Estimation} show mode-covering limits the perceptual quality of reconstructions. To reiterate, this is because mode-covering produces \textit{unrealistic} samples.

The question then arises: can we do better and produce \textit{realistic} reconstructions using only a 1 bit encoding?

\subsection{Quantized Hierarchies}
\label{sec:prob_quant_hier_motivation}
We now take the pretrained 4-code VQ-VAE that produces the red trace in Figure~\ref{fig:toy_single_layer_vqvae}. We term this VQ-VAE Layer 1. We then train a \textit{new} 2-code VQ-VAE, which we term Layer 2, to compress and reconstruct the encodings produced by Layer 1. The resulting system is a \textit{quantized hierarchy}.

We can then compress and reconstruct datapoints sampled from the original distribution using the whole quantized hierarchy, as shown in Algorithm \ref{alg:quantized_hierarchy_lossy_compression}. Algorithm \ref{alg:quantized_hierarchy_lossy_compression} is a simplification of the Hierarchical Quantized Autoencoder (HQA) described in Section~\ref{section:hqm-methods}.

\begin{algorithm}[h!]
\caption{Lossy Compression Pseudo-code Using A Quantized Hierarchy}
\label{alg:quantized_hierarchy_lossy_compression}
\begin{algorithmic}[1]
\State $x$: Datapoint to be compressed
\State $e_1 \gets \text{Encoder}_1(x)$ \Comment{Encode using Layer 1}
\State $e_2 \gets \text{Encoder}_2(e_1)$ \Comment{Encode using Layer 2}
\State $q_2 \gets \text{Quantize}(e_2)$
\State Transmit $q_2$ \Comment{$q_2$ is the final encoding}
\State $\hat{e}_1 \gets \text{Decoder}_2(q_2)$ \Comment{Decode using Layer 2; this will mode-cover Layer 1's latent space}
\State $\hat{q}_1 \gets \text{Quantize}(\hat{e}_1)$ \Comment{Resolve mode-covering in the latent space}
\State $\hat{x} \gets \text{Decoder}_1(\hat{q}_1)$ \Comment{Decode using Layer 1}
\State $\hat{x}$: Lossy reconstruction of $x$
\end{algorithmic}
\end{algorithm}

For VQ-VAE, each codeword is represented as a vector in a continuous latent space. If we consider the points in the latent space of Layer 1 that are actually used for encodings, there are only 4 points that are used: the locations of the 4 codewords. In other words, the distribution over the latent space of Layer 1 contains 4 modes.

Layer 2 is used to compress and reconstruct points from the latent space of Layer 1. As the latent space of Layer 1 contains 4 modes but Layer 2 only uses 2 codewords, Layer 2 will mode-cover for the same reasons described above. However, this mode-covering is now over the \textit{latent} space of Layer 1 and not over the \textit{input} space of the original distribution.

This mode-covering can now be resolved through quantization. The decode from Layer 2 can be quantized to a code in Layer 1's latent space (i.e quantized to a mode). Therefore, the reconstructions of Layer 1's latent space, and hence the final reconstruction, are more likely to be \textit{realistic}.

VQ-VAE uses a deterministic quantization procedure which always quantizes to the code that is geometrically closest to the input embedding. We can use the quantized hierarchy introduced, along with deterministic quantization, to reconstruct samples from the original distribution. The result is shown by the red trace in Figure \ref{fig:toy_deterministic}. For the reasons outlined above, no mode-covering behaviour is observed and all reconstructions are \textit{realistic}. However, mode-dropping is now occurring.

\subsection{Stochastic Quantization}
\label{sec:stoch_motivation}

If a stochastic quantization scheme is introduced (c.f. Section~\ref{section:stoch_post}) then this mode-dropping behaviour can also be resolved. Figure~\ref{fig:toy_stochastic} shows the result of using the quantized hierarchy, now with stochastic quantization. No mode-dropping or mode-covering behaviour is present. Note that the quantized hierarchy uses 1 bit encodings, the same size as the encoding of the 2-code VQ-VAE that failed to model the distribution (c.f. Figure \ref{fig:toy_single_layer_vqvae}). This result shows that, under a given information bottleneck, probabilistic quantized hierarchies allow for fundamentally different density modelling behaviour than equivalent single layer systems. Furthermore, unlike deterministic compression, there is no single decoded data; there are now many possible decodes.

Therefore, we propose that probabilistic quantized hierarchies can mitigate the \textit{unrealistic} reconstructions produced by likelihood-based systems for the following reasons:
\begin{itemize}
    \itemsep0em
    \item \textbf{Hierarchy:} By choosing to model a distribution using a hierarchical latent space of increasingly compressed representations, mode-covering behaviour in the input space can be exchanged for mode-covering behaviour in the latent space. This also acts as a good meta-prior to match the hierarchical structure of natural data \cite{lazaro2016hierarchical}.
    
    \item \textbf{Quantization:} Quantization allows for the resolution of mode-covering behaviour in latent space, encouraging \textit{realistic} reconstructions that fall in regions of high density in the input space.
    
    \item \textbf{Stochastic Quantization:} If quantization is performed deterministically then diversity of reconstructions is sacrificed. By quantizing stochastically, mode-dropping behaviour can be mitigated. In addition, this introduces the stochasticity typically required for low-rate lossy compression in a natural manner.
\end{itemize}

\section{Method}
\label{method}

\subsection{Stochastic Posterior}
\label{section:stoch_post}
We depart from the deterministic posterior of VQ-VAE and instead use the stochastic posterior introduced by \citet{ConRelax}:
\begin{equation}
    q(z=k|x) \propto \exp-||z_e(x) - e_k||_2^2 \: .
    \label{eqn:stochastic_posterior}
\end{equation}
Quantization can then be performed by sampling from $q(z=k|x)$. At train-time, a differentiable sample can be obtained from this posterior using the Gumbel Softmax relaxtion \cite{Gumbel,Concrete}. While training HQA, we linearly decay the Gumbel Softmax temperature to $0$ so the soft quantization operation closely resembles hard quantization, which is required when compressing to a fixed rate. At test-time we simply take a sample from Equation~\ref{eqn:stochastic_posterior}.

Crucially, under this formulation of the posterior, $z_e(x)$ (henceforth $z_e$) must be positioned well relative to \emph{all} codes in the latent space, not just the nearest code \cite{vq_wu}. As $z_e$ implicitly defines a distribution over all codes, it carries more information about $x$ than a single quantized latent sampled from $q(z|x)$. This is exploited by the HQA hierarchy, as discussed below.

\subsection{Training Objective}
\subsubsection{Single Layer}
\label{section:hqm-methods}
In a single layer model, the encoder generates a posterior $q = q(z|x)$ over the codes given by Equation~\ref{eqn:stochastic_posterior}. To calculate a reconstruction loss we sample from this posterior and decode. Additionally, we augment this with two loss terms that depend on $q$: 
\begin{equation}
    \label{eqn:our_loss}
    \mathcal{L} =  \underbrace{-\log p(x|z=k)}_{\text{reconstruction loss}} - \underbrace{\mathcal{H}[q(z|x)]}_{\text{entropy}} + \underbrace{\mathbb{E}_{q(z|x)}||z_e(x) - e_z||_2^2}_{\text{probabilistic commitment loss}} \: .
\end{equation}
This objective is the sum of the \textit{reconstruction loss} as in a normal VQ-VAE (Equation~\ref{eqn:vqvaeloss}), the \textit{entropy} of $q$, and a term similar to the \textit{codebook/commitment loss} in Equation~\ref{eqn:vqvaeloss} but instead taken over all codes, weighted by their probability under $q$. The objective $\mathcal{L}$ resembles placing a Gaussian Mixture Model (GMM) prior over the latent space and calculating the Evidence Lower BOund (ELBO), which we derive in Appendix~\ref{sec:prop_vqvae}.

\subsubsection{Multiple Layers}
When training higher layers of HQA, we take take the reconstruction target to be $z_e$ from the previous layer. This novel choice of reconstruction target is motivated by noting that the embedding of $z_e$ implicitly represents a distribution over codes. By training higher layers to minimize the MSE between $z_e$ from the layer below and an estimate $\hat{z}_e$, the higher layer learns to reconstruct a full distribution over code indices, not just a sample from this distribution. Empirically, the results in Section~\ref{sec:mnist} show this leads to gains in reconstruction quality.

In this way, a higher level VQ-VAE can be thought of as reconstructing the full posterior of the layer below, as opposed to a sample from this posterior (as in \citet{HAMS}). The predicted $\hat{z}_e$ is used to estimate the posterior of the layer below using Equation~\ref{eqn:our_loss}, from which we can easily sample to perform stochastic quantization, as motivated in Section \ref{motivation}.

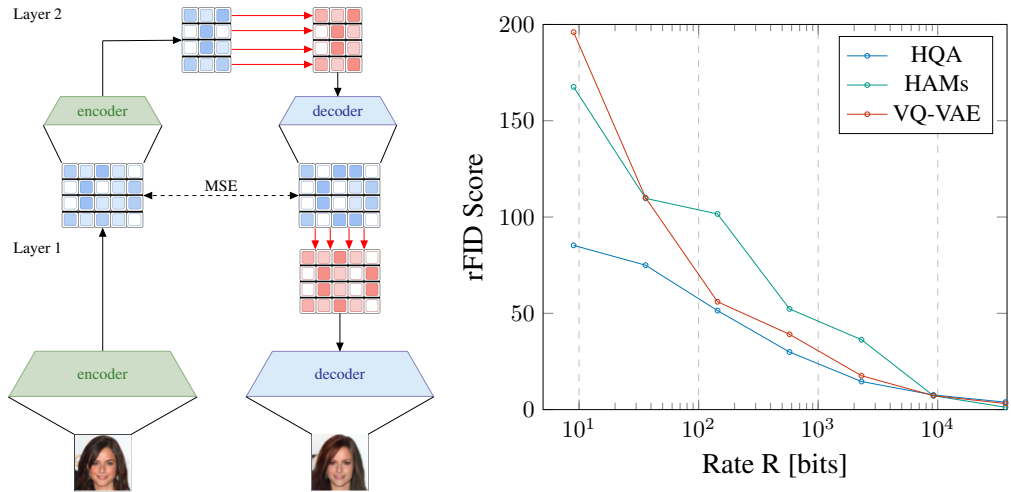
\begin{figure}[!h]
    \vspace{-1em}
    \centering
    \begin{subfigure}[c]{.35\textwidth}
        \centering
        \scalebox{0.5}{\input{tikzdiagrams/method-diagram.tex}}
    \end{subfigure}
    ~
    \hspace{2em}
    \begin{subfigure}[c]{0.45\textwidth}
        \centering
        \begin{tikzpicture}[xscale=0.9,yscale=0.9]
        \begin{axis}[
            xlabel={\large Rate R [bits]},
            ylabel={\large rFID Score},
            xmode=log,
            xmin=5, xmax=38000,
            ymin=0, ymax=200,
            yticklabel style={/pgf/number format/fixed},
            xtick={10,100,1000,10000},
            ytick={0,50,100,150,200},
            legend pos=north east,
            xmajorgrids=true,
            grid style=dashed,
        ]
        
        \addplot[,color=mycolour1,mark=o,mark size=1pt,]
            coordinates {
                (36864.0,3.88)(9216,7.59)(2304,14.6)(576,29.9)(144,51.4)(36,75)(9,85.3)
            };
        \addplot[,color=mycolour2,mark=o,mark size=1pt,]
            coordinates {
               (36864.0,1.24)(9216,7.2)(2304,36.3)(576,52.3)(144,101.6)(36,109.7)(9,167.6)    };
        \addplot[,color=mycolour3,mark=o,mark size=1pt,]
            coordinates {
            (36864.0,3.16)(9216,7.24)(2304,17.6)(576,39.1)(144,56)(36,110)(9,196)
            };
            \legend{HQA, HAMs, VQ-VAE}
        
        \end{axis}
        \end{tikzpicture}

        \label{fig:celeba_fid}  
    \end{subfigure}
    \hspace{2em}
    \vspace{-0.5em}
    \caption{
      \textbf{Left}:  System diagram of training the second layer of the HQA. Images are encoded into a continuous latent vector by Layer 1 before being encoded further by Layer 2. This representation is then quantized according to the stochastic posterior given by the red arrows, and then decoded by Layer 2. If training, an MSE loss is taken with this output and the input to the Layer 2 encoder. If performing a full reconstruction, the representation is quantized and then decoded by Layer 1.
      \textbf{Right}: Plot of rate against reconstruction FID (rFID) for compressing and reconstructing CelebA test examples.
    }
    \label{fig:systemdiagram}
    \vspace{-0.5em}
\end{figure}

The Markovian latent structure of HQA - where each latent space is independent given the previous layer - allows us to train each layer sequentially in a greedy manner as shown in Figure~\ref{fig:systemdiagram} (left). This leads to lower memory footprints and increased flexibility as we are able to ensure the performance of each layer before moving onto the next. Appendix~\ref{sec:algorithm} describes algorithm in full.

\subsection{Codebook Optimization}
\label{section:codeopt}
The loss given by Equation~\ref{eqn:our_loss}, in combination with the use of the Gumbel-Softmax, allows for the code embeddings to be learnt directly without resorting to moving average methods. This introduces a new pathology where codes that are assigned low probability under $q(z|x)$ for all $x$ receive low magnitude gradients and become unused. During training, we reinitialise these unused codes near codes of high usage. This results in significantly higher effective rates. Code resetting mirrors prior work in online GMM training~\cite{fastGMM,EfficientGMM} and over-parameterized latent spaces \cite{overparamEM}.

\section{Experiments}
\label{setup}

\subsection{CelebA}

\begingroup
\setlength{\tabcolsep}{1pt} %

\begin{table*}[ht!]
\vspace{-1em}
\caption{Reconstructed CelebA test-set images at different levels of compression, with number of transmitted bits}
\vspace{0.5em}
\small

\centering
\begin{tabular}{c cccccccc}
\toprule
\multirow{2}{*}{System} & Original & 2.7x & 11x & 43x & 171x & 683x & 2,731x & 10,923x \\
 & 98,304 & 36,864 & 9,216 & 2,304 & 576 & 144 & 36 & 9 bits  \\ \midrule
HQA  & \includegraphics[scale=0.25, align=c]{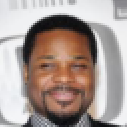} &
\includegraphics[scale=0.25, align=c]{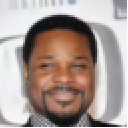} & 
\includegraphics[scale=0.25, align=c]{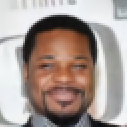} & 
\includegraphics[scale=0.25, align=c]{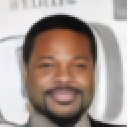} & 
\includegraphics[scale=0.25, align=c]{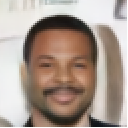} & 
\includegraphics[scale=0.25, align=c]{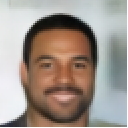} & 
\includegraphics[scale=0.25, align=c]{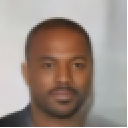} & 
\includegraphics[scale=0.25, align=c]{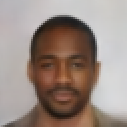} \\
HAMs & \includegraphics[scale=0.25, align=c]{celeba_results/side_by_sides/celeba_val_img1_orig} &
\includegraphics[scale=0.25, align=c]{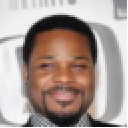} &
\includegraphics[scale=0.25, align=c]{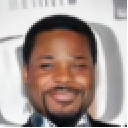} &
\includegraphics[scale=0.25, align=c]{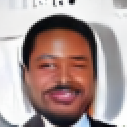} &
\includegraphics[scale=0.25, align=c]{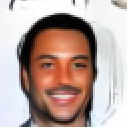} & 
\includegraphics[scale=0.25, align=c]{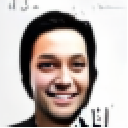} &
\includegraphics[scale=0.25, align=c]{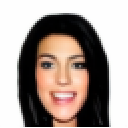} & 
\includegraphics[scale=0.25, align=c]{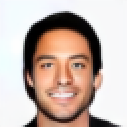} \\
VQ-VAE & \includegraphics[scale=0.25, align=c]{celeba_results/side_by_sides/celeba_val_img1_orig} &
\includegraphics[scale=0.25, align=c]{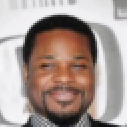} & 
\includegraphics[scale=0.25, align=c]{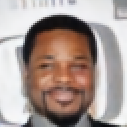} & 
\includegraphics[scale=0.25, align=c]{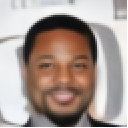} & 
\includegraphics[scale=0.25, align=c]{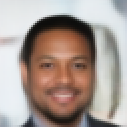} & 
\includegraphics[scale=0.25, align=c]{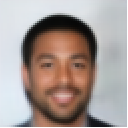} & 
\includegraphics[scale=0.25, align=c]{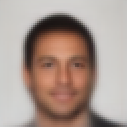} & 
\includegraphics[scale=0.25, align=c]{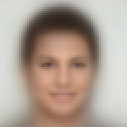} \\
\midrule
HQA &  \includegraphics[scale=0.25, align=c]{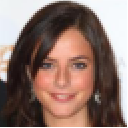} &
\includegraphics[scale=0.25, align=c]{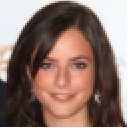} & 
\includegraphics[scale=0.25, align=c]{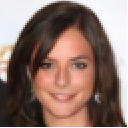} &
\includegraphics[scale=0.25, align=c]{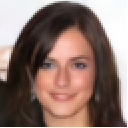} & 
\includegraphics[scale=0.25, align=c]{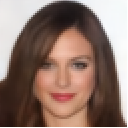} & 
\includegraphics[scale=0.25, align=c]{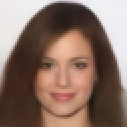} & 
\includegraphics[scale=0.25, align=c]{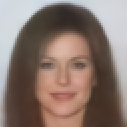} & 
\includegraphics[scale=0.25, align=c]{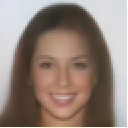} \\
HAMs & \includegraphics[scale=0.25, align=c]{celeba_results/side_by_sides/celeba_val_img2_orig} &
\includegraphics[scale=0.25, align=c]{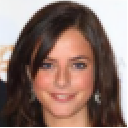} & 
\includegraphics[scale=0.25, align=c]{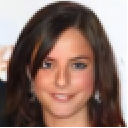} & 
\includegraphics[scale=0.25, align=c]{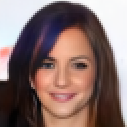} &
\includegraphics[scale=0.25, align=c]{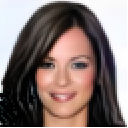} &
\includegraphics[scale=0.25, align=c]{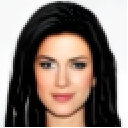} & 
\includegraphics[scale=0.25, align=c]{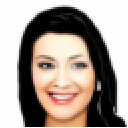} &
\includegraphics[scale=0.25, align=c]{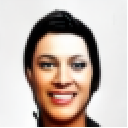} \\
VQ-VAE & \includegraphics[scale=0.25, align=c]{celeba_results/side_by_sides/celeba_val_img2_orig} &
\includegraphics[scale=0.25, align=c]{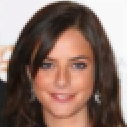} & 
\includegraphics[scale=0.25, align=c]{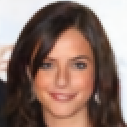} & 
\includegraphics[scale=0.25, align=c]{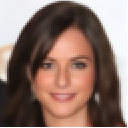} & 
\includegraphics[scale=0.25, align=c]{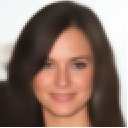} & 
\includegraphics[scale=0.25, align=c]{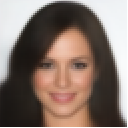} & 
\includegraphics[scale=0.25, align=c]{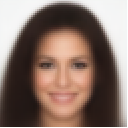} & 
\includegraphics[scale=0.25, align=c]{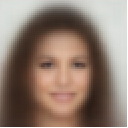} \\
\bottomrule
\end{tabular}

\label{tab:celeba}
\end{table*}
\endgroup

To show the scalability of HQA and the compression rates it can achieve on natural images, we train on the CelebA dataset \cite{CelebA} at a 64x64 resolution. The resulting system is a 7-layer HQA, where the final latent space of 512 codes has size $1 \times 1$ due to downsampling by 2 at each layer. The architecture of each layer is detailed in Appendix~\ref{sec:arch_hparams}.

For comparison, we also train 7 different VQ-VAE systems. Each VQ-VAE has the same compression ratio and approximate parameter count as its HQA equivalent. We also compare against the hierarchical quantized system introduced by HAMs, since their system also can be used for low-rate compression with simple feedforward decoders (c.f. discussion in Section~\ref{sec:vae_hierarchy_relatedwork}). As with the VQ-VAE baselines, each HAMs layer has the same compression ratio as its HQA equivalent. Table~\ref{tab:celeba} shows reconstructions of two different images from the test set for each layer of HQA, as well as the reconstructions from the VQ-VAE and HAMs baselines.

Qualitatively, the HQA reconstructions display higher perceptual quality than both VQ-VAE and HAMs at all compression rates, with the difference becoming more exaggerated as the compression becomes more extreme. The high-level semantic features of the input image are also better preserved with HQA than with the baselines, even when the reconstructions are very different from the original in pixel space. For a quantitative comparison, we evaluate the test set reconstruction Fr\'echlet Inception Distance (rFID) for each system. Figure~{\ref{fig:systemdiagram}} (right) shows that HQA achieves better rFIDs than both VQ-VAE and HAMs and, as with the qualitative comparison, the difference becomes more exaggerated at low rates. We note the well known issues with relative comparison between likelihood-based models and adversarially trained models when using rFID \cite{ravuri2019classification}, and therefore only look to compare HQA with likelihood-based baselines.

\subsection{MNIST}
\label{sec:mnist}

We performed an ablation study on MNIST \cite{mnist} with the data rescaled to 32x32. In addition to measuring distortion and rFID, we evaluated how well each system was preserving the semantic content of each image by using a pre-trained MNIST classifier to classify the resulting reconstructions.

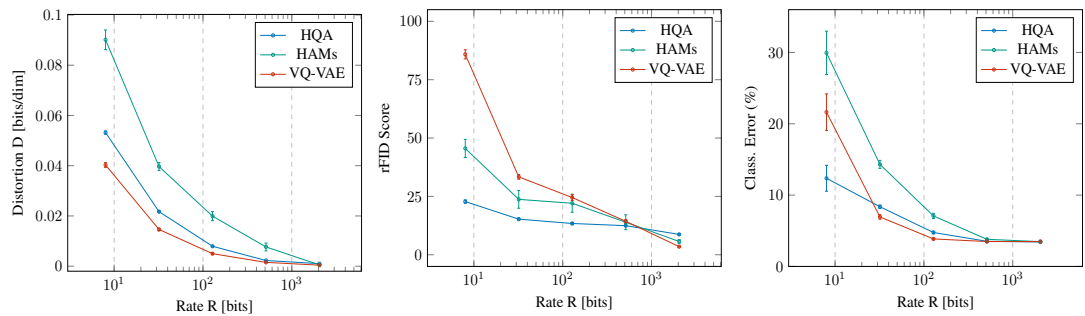
\begin{figure*}[h]
    \begin{subfigure}[l]{0.25\textwidth}
        \begin{tikzpicture}[xscale=0.57,yscale=0.6]
        \begin{axis}[
            xlabel={Rate R [bits]},
            ylabel={Distortion D [bits/dim]},
            ylabel near ticks,
            xmode=log,
            ymin=-0.002, ymax=0.1002,
            xmin=3, xmax=6100,
            yticklabel style={/pgf/number format/fixed},
            ytick={0,0.020,0.040,0.060,0.080,0.100},
            xtick={10,100,1000,10000},
            legend pos=north east,
            xmajorgrids=true,
            grid style=dashed,
        ]
        \addplot+[color=mycolour1,mark=o,mark size=1pt,] plot [error bars/.cd,y dir=both, y explicit,]
            coordinates {
                (2048.0,0.001006411267403326) +- (0,2.9079171083177658e-05)
                (512.0,0.002265735716698021) +- (0,4.93421731184553e-05)
                (128.0,0.007936912284537762) +- (0,0.00017594461964248176)
                (32.0,0.021720726032848803) +- (0,0.0003458508902526826)
                (8.0,0.05318145127446905) +- (0,0.0005933246693197752)
            };
        \addplot+[color=mycolour2,mark=o,mark size=1pt,] plot [error bars/.cd,y dir=both, y explicit,]
            coordinates {
                (2048.0,0.0005255426467625918) +- (0,6.398481215555307e-05)
                (512.0,0.007663635767273393) +- (0,0.001547007343965844)
                (128.0,0.019925941618112043) +- (0,0.001811837333090433)
                (32.0,0.03966354457843531) +- (0,0.0015627099183824602)
                (8.0,0.0901077096225044) +- (0,0.003908269556968918)
              };
        \addplot+[color=mycolour3,mark=o,mark size=1pt,] plot [error bars/.cd,y dir=both, y explicit,]
            coordinates {
                (2048.0,0.00041335128611561434) +- (0,2.3097347055205916e-05)
                (512.0,0.001520898798828697) +-     (0,0.000104568966443733)
                (128.0,0.00498923244701203) +- (0,0.0003316210721342356)
                (32.0,0.014618005982759636) +- (0,0.0004905543724891704)
                (8.0,0.04028539200380095) +- (0,0.0009355666803056496)
            };
            \legend{HQA, HAMs, VQ-VAE}
        \end{axis}
        \end{tikzpicture}
    \end{subfigure}
    \hspace{12mm}
    \begin{subfigure}[l]{0.25\textwidth}
        \begin{tikzpicture}[xscale=0.57,yscale=0.6]
        \begin{axis}[
            xlabel={Rate R [bits]},
            ylabel={rFID Score},
            ylabel near ticks,
            xmode=log,
            xmin=3, xmax=6100,
            ymin=-5, ymax=105,
            xtick={10,100,1000,10000},
            yticklabel style={/pgf/number format/fixed},
            ytick={0,25,50,75,100,200,300},
            legend pos=north east,
            xmajorgrids=true,
            grid style=dashed,
        ]
        
        \addplot+[color=mycolour1,mark=o,mark size=1pt,] plot [error bars/.cd,y dir=both, y explicit,]
            coordinates {
                (2048.0,8.72298592126647) +- (0,0.30543614926659407)
                (512.0,12.461351356487521) +- (0,0.26977174641327434)
                (128.0,13.399912095161273) +-         (0,0.4836518754149487)
                (32.0,15.259035720523022) +- (0,0.4902534385725001)
                (8.0,22.785450519762275) +- (0,0.708806851727058)
            };
        \addplot+[color=mycolour2,mark=o,mark size=1pt,] plot [error bars/.cd,y dir=both, y explicit,]
            coordinates {
                (2048.0,5.690082432726484) +- (0,0.783969129236414)
                (512.0,13.945582935840317) +- (0,3.1829070160862787)
                (128.0,22.062132571524195) +- (0,3.87355122976046)
                (32.0,23.733071861230066) +- (0,3.805581705255765)
                (8.0,45.5579206972327) +- (0,3.871331339327621)
            };
        \addplot+[,color=mycolour3,mark=o,mark size=1pt,] plot [error bars/.cd,y dir=both, y explicit,]
            coordinates {
                (2048.0,3.4809279876012225) +- (0,0.24251297973445032)
                (512.0,14.3256856269498) +-             (0,0.6819633512498714)
                (128.0,24.507091260643758) +- (0,1.2555867886624281)
                (32.0,33.352211268586714) +- (0,0.9939611062753853)
                (8.0,85.85506995998428) +- (0,1.9637878021286574)
            };
            \legend{HQA, HAMs, VQ-VAE}
        
        \end{axis}
        \end{tikzpicture}
    \end{subfigure}
    \hspace{12mm}
    \begin{subfigure}[l]{0.25\textwidth}
        \begin{tikzpicture}[xscale=0.57,yscale=0.6]
        \begin{axis}[
            xlabel={Rate R [bits]},
            ylabel={Class. Error (\%)},
            ylabel near ticks,
            xmode=log,
            ymin=0, ymax=36,
            xmin=3, xmax=6100,
            yticklabel style={/pgf/number format/fixed},
            ytick={0,10,20,30,40,75,100,200,300},
            xtick={10,100,1000},
            legend pos=north east,
            xmajorgrids=true,
            grid style=dashed,
        ]
        
        \addplot+[color=mycolour1,mark=o,mark size=1pt,] plot [error bars/.cd,y dir=both, y explicit,]
            coordinates {
                (2048.0,3.4229999999999974) +- (0,0.09243431484032447)
                (512.0,3.525) +- (0,0.06852996424922503)
                (128.0,4.756999999999996) +- (0,0.16856113953103088)
                (32.0,8.366) +- (0,0.23326799831952894)
                (8.0,12.361) +- (0,1.8188408206987194)
            };
        \addplot+[color=mycolour2,mark=o,mark size=1pt,] plot [error bars/.cd,y dir=both, y explicit,]
            coordinates {
               (2048.0,3.474000000000001) +- (0,0.09223252701731692)
                (512.0,3.8049999999999997) +- (0,0.11942336454815007)
                (128.0,7.080000000000001) +- (0,0.3724928305350328)
                (32.0,14.306999999999999) +- (0,0.5590025146276182)
                (8.0,29.943) +- (0,3.0388269442270004)
            };
       \addplot+[color=mycolour3,mark=o,mark size=1pt,] plot [error bars/.cd,y dir=both, y explicit,]
            coordinates {
                (2048.0,3.4909999999999983) +- (0,0.10643539937445547)
                (512.0,3.490000000000002) +- (0,0.09346549309771979)
                (128.0,3.8579999999999983) +- (0,0.11323305241845406)
                (32.0,6.939) +- (0,0.3151939616172868)
                (8.0,21.631999999999994) +- (0,2.557611202227579)
            };
    
        \legend{HQA, HAMs, VQ-VAE}
        
        \end{axis}
        \end{tikzpicture}
    \end{subfigure}
    \caption{Plots of rate against distortion, reconstruction FID (rFID) and classification error for compressing and then reconstructing MNIST test examples.  Error bars are 95\% confidence intervals based on 10 runs with different training seeds.}
    \label{fig:mnistrplot}
    \vspace{-1em}
\end{figure*}

\begin{table*}[h]
\centering
\small
\caption{Distortion (MSE), reconstruction FID (rFID) and Classification Error scores for ablated systems, after compressing MNIST 10k test samples into an 8-bit 1x1 latent space then reconstructing. `GS' covers introducing Gumbel Softmax and code resetting. `MSE' means using Mean Squared Error loss on all layers. Errors represent a 95\% confidence interval based on 10 runs.}
\label{tab:mnist_ablation}
\vspace{0.5em}
\begin{tabular}{lcccm{2cm}}
\toprule
System & Distortion $\downarrow$ & rFID Score $\downarrow$ & Class. Error (\%) $\downarrow$ & Reconstructions \\
\midrule
No Compression      & $0.000 \pm 0.000$         & $0.0 \pm 0.0$         & $3.13 \pm 0.00$ & \vspace{-0.1em} \includegraphics[trim={3mm 0 4mm 0}, clip,width=0.13\textwidth]{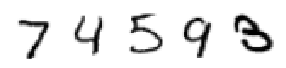} \\
\hline
VQ-VAE              & $\bm{0.040 \pm 0.001}$    & $85.9 \pm 2.0$        & $21.6 \pm 2.56$ & \vspace{0.1em} \includegraphics[trim={3mm 0 4mm 0}, clip,width=0.13\textwidth]{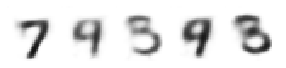} \\
+ hierarchy (HAMs)  & $0.090 \pm 0.004$         & $45.6 \pm 3.9$        & $29.9 \pm 3.04$ & \includegraphics[trim={3mm 0 4mm 0}, clip,width=0.13\textwidth]{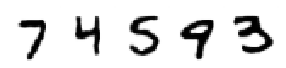} \\
\hline
HAMs + GS           & $0.108 \pm 0.009$         & $38.6 \pm 3.1$        & $51.1 \pm 6.28$  & \vspace{0.1em} \includegraphics[trim={3mm 0 4mm 0}, clip,width=0.13\textwidth]{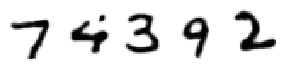} \\
HAMs + MSE          & $0.052 \pm 0.0004$        & $36.0 \pm 1.0$        & $11.1 \pm 0.46$ & \includegraphics[trim={3mm 0 4mm 0}, clip,width=0.13\textwidth]{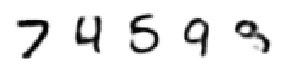} \\
\hline
HAMs + GS + MSE     & $0.054 \pm 0.0003$        & $\bm{21.0 \pm 1.0}$   & $\bm{10.6 \pm 0.93}$ & \vspace{0.1em} \includegraphics[trim={3mm 0 4mm 0}, clip,width=0.13\textwidth]{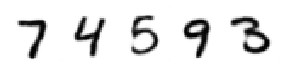} \\
+ probabilistic loss (\textbf{HQA}) & $0.053 \pm 0.0006$        & $22.8 \pm 0.7$            & $12.4 \pm 1.82$ & \includegraphics[trim={3mm 0 4mm 0}, clip, width=0.13\textwidth]{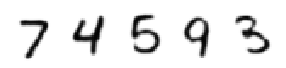} \\
\end{tabular}

\vspace{-1em}
\end{table*}

We trained five layers, each compressing the original images by a factor of 2 in each dimension, such that the final layer compressed to a latent space of size 1x1. For VQ-VAE we trained to a 1x1 latent space directly. We control for the number of parameters ($\sim$1M) in each system, training each with codebook size 256 and dimension 64.

Table~\ref{tab:mnist_ablation} and Figure~\ref{fig:mnistrplot} both show that HQA has superior rate-perception performance (as approximated by rFID) at low rates than the other baselines. The trade-off between rate-perception and rate-distortion performance described by \citet{blau2019rethinking} is clearly visible, resulting in HQA displaying worse distortions but better rFID scores. Furthermore, the classification accuracy results show that, at extreme rates, HQA maintains more semantic content from the originals when compared to the other methods. 

Furthermore, the ablation study in Table~\ref{tab:mnist_ablation} shows that, although the Gumbel-Softmax (GS) and MSE loss show improved performance when used individually, it is the combination of both that leads to the largest gain in performance, suggesting the benefits are orthogonal. Notably, HQA is the only system to give both good rFID and classification scores across all rates, the largest difference being at extreme compression rates. We note that the probabilistic loss of HQA hinders performance under the MNIST task. However, we empirically found that the probabilistic loss was essential to ensure stability of HQA when training on more complex datasets such as CelebA.

\begin{table}[h]
\vspace{-0.5em}
\centering
\small
\tabcolsep=0.11cm
\caption{Linear interpolations of encoder output $z_e$ in the 8 bit 1x1 latent space. The far left and right images are originals. Others are decoded from the interpolated quantized encoder output $z_q$.}
\label{tab:interp-models}
\vspace{0.5em}
\begin{tabular}{m{1.2cm}||m{0.45cm}|m{4.7cm}|m{0.45cm}}
HQA	&	 \includegraphics[trim={0 2mm 0 2mm}, clip,height=0.4cm,keepaspectratio]{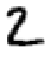} 	&   \includegraphics[height=0.59cm,keepaspectratio]{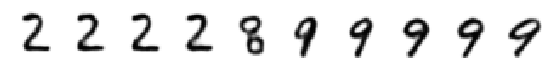}  &  \includegraphics[trim={0 1mm 0 2mm}, clip,height=0.4cm,keepaspectratio]{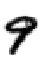} \\ 
HAMs	&	 \includegraphics[trim={0 2mm 0 2mm}, clip,height=0.4cm,keepaspectratio]{mnist_results/interpolations/2-orig.png} 	&   \includegraphics[height=0.5cm,keepaspectratio]{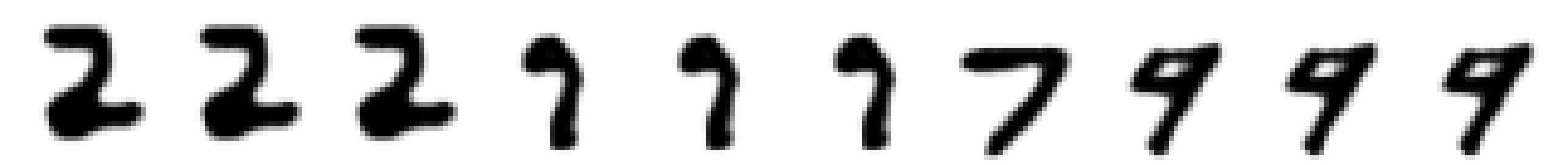}  &  \includegraphics[trim={0 1mm 0 2mm}, clip,height=0.4cm,keepaspectratio]{mnist_results/interpolations/9-orig.png} \\
VQ-VAE &	 \includegraphics[trim={0 2mm 0 2mm}, clip,height=0.4cm,keepaspectratio]{mnist_results/interpolations/2-orig.png} 	&   \includegraphics[height=0.5cm,keepaspectratio]{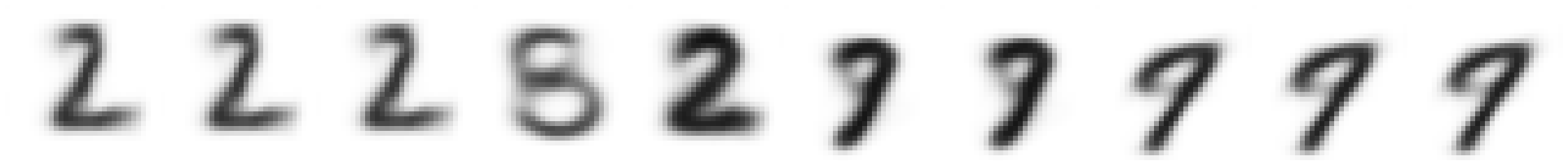}  &  \includegraphics[trim={0 1mm 0 2mm}, clip,height=0.4cm,keepaspectratio]{mnist_results/interpolations/9-orig.png} \\ 
\end{tabular}
\vspace{-0.5em}
\end{table}

Linear interpolations in Table~\ref{tab:interp-models} show that HQA has more dense support for coherent representations across its latent space than HAMs or VQ-VAE. Intermediate images for HQA are sharp and crisply represent digits, never deforming into unrealistic shapes. The same behaviour is observed for faces in the CelebA dataset, as shown in Table~\ref{tab:interp-celeba}. Additional results can be found in Appendix~\ref{sec:additional_results}.

\section{Conclusion}
In this work, we introduce the `Hierarchical Quantized Autoencoders', a promising method for training hierarchical VQ-VAEs under low-rate lossy compression. HQA introduces a new objective and is a naturally stochastic system. By incorporating a variety of additional improvements, we show HQA outperforms equivalent VQ-VAE architectures when reconstructing on the CelebA and MNIST datasets under extreme compression.

\clearpage
\section*{Broader Impact}

It is estimated that streaming of digital media accounts for 70\% of today's internet traffic \cite{li2019neural}, and this is reflected by the increasing importance of high quality compact representations in the big visual data era \cite{Ma_2020}. Our research takes steps towards addressing this issue by providing a scalable architecture for semantically meaningful compression, at rates unachievable by traditional algorithms. 

As well as the economic advantages of low-rate compression, there is the benefit of reduced energy and resources required for transmission and storage of smaller data, although this must be traded off against the currently higher computational cost of encoding/decoding.

Like most image based research, HQA has broader implications related to computer vision applications and the ethics surrounding them. As these are detailed by \citet{Lauronen2017EthicalII} we instead choose to focus more directly on the potential consequences of our cited objective: to produce realistic and semantically consistent compressed images at low bitrates. 

Whilst we observe empirically that the hierarchy of concepts retained by the HQA model can relate to a human idea of semantic importance, we do not control for this explicitly, which could have negative repercussions. 

For example, in the case of human imagery it is possible for decoded characteristics related to ethnicity or gender to be misrepresentative of the original, a scenario which may be exacerbated by a biased training set. In a more general sense, it is possible that mission critical details could be removed or modified, and whilst this is symptomatic of all low bitrate lossy compressions schemes, the realism of the output could lead to an misguided interpretation which would traditionally be offset by the appearance of artifacts or a lower resolution output. 

An interesting future research direction could be to alleviate this issue by conditioning the model on semantic labels as demonstrated by \citet{agustsson2019generative}.

Further to this, the stochastic nature of our decodes means that the sender of an image has no way of knowing exactly what image the receiver will view and indeed different receivers of the same transmitted image will see different outputs. To a degree, viewers of media are used to this (for example where technologies automatically increase / reduce resolution according to available bandwidth), however methods such as ours have the potential to vary images in terms of higher level content as well as fine grained detail. This makes quality control, for example, problematic and use cases sensitive to this would need to do careful further investigation before using techniques such as ours. For other use cases however, such as artistic media, having a built in method for variable user experience may actually provide an interesting avenue for creative exploration.

\bibliography{paper}
\bibliographystyle{abbrvnat}

\appendix
\onecolumn

\section{Additional HQA Results}\label{sec:additional_results}

\begin{table*}[htbp!]

\caption{Additional CelebA interpolations of the HQA encoder output $z_e$ in the 9 bit 8x8 latent space. Compression is from 98,304 to 576 bits (171x compression).}

\tabcolsep=0.11cm
\small
\centering
\begin{tabular}{c|c|c}
\includegraphics[height=1.4cm,keepaspectratio]{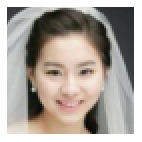} 	&   \includegraphics[height=1.4cm,keepaspectratio]{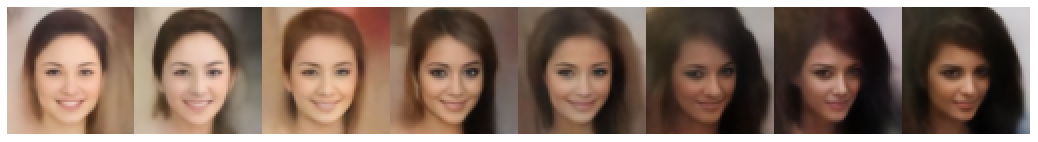}  &  \includegraphics[height=1.4cm,keepaspectratio]{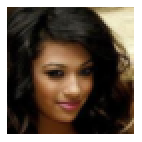} 	\\
\includegraphics[height=1.4cm,keepaspectratio]{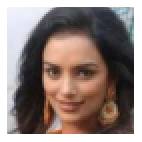} 	&   \includegraphics[height=1.4cm,keepaspectratio]{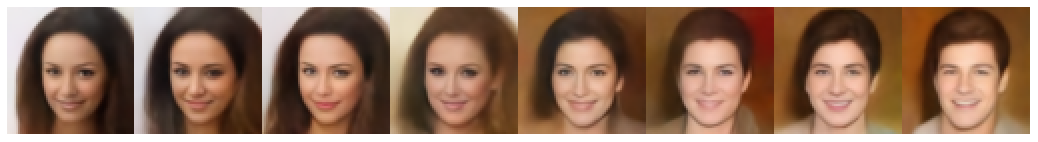}  &  \includegraphics[height=1.4cm,keepaspectratio]{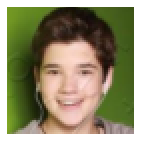}  \\	\includegraphics[height=1.4cm,keepaspectratio]{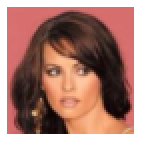} 	&   \includegraphics[height=1.4cm,keepaspectratio]{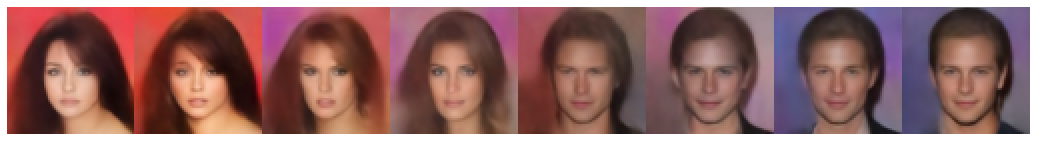}  &  \includegraphics[height=1.4cm,keepaspectratio]{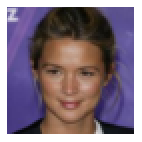} \\ 
\includegraphics[height=1.4cm,keepaspectratio]{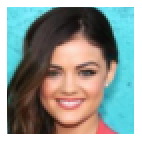} 	&   \includegraphics[height=1.4cm,keepaspectratio]{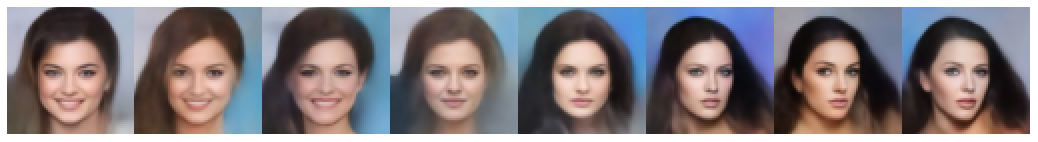}  &  \includegraphics[height=1.4cm,keepaspectratio]{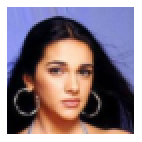} \\ 
\includegraphics[height=1.4cm,keepaspectratio]{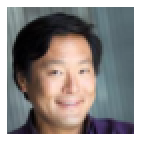} 	&   \includegraphics[height=1.4cm,keepaspectratio]{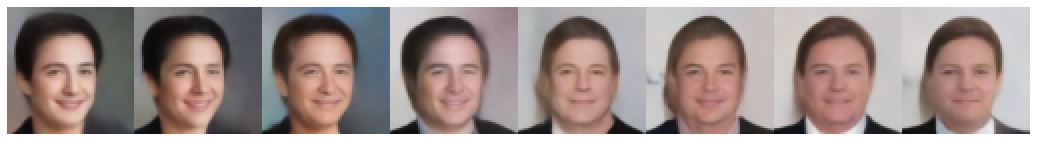}  &  \includegraphics[height=1.4cm,keepaspectratio]{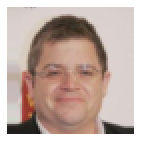} \\ 
\includegraphics[height=1.4cm,keepaspectratio]{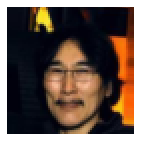} 	&   \includegraphics[height=1.4cm,keepaspectratio]{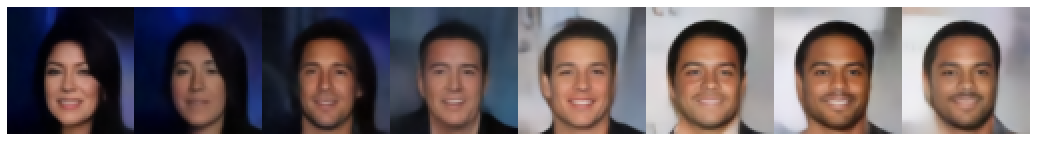}  &  \includegraphics[height=1.4cm,keepaspectratio]{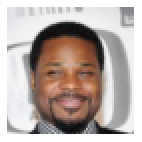}  \\

\end{tabular}
\end{table*}

\begingroup
\setlength{\tabcolsep}{1pt} %
\centering
\begin{table*}[htbp!]
\vspace{-1em}
\caption{CelebA reconstruction diversity when performing stochastic decodes from the 9 bit 4x4 latent space. Compression is from 98,304 to 144 bits (683x compression).}
\vspace{0.5em}
\small

\centering
\begin{tabular}{c|cccccc}
\toprule
Original & \multicolumn{6}{c}{Stochastic Reconstructions} \\ \midrule
\includegraphics[scale=0.3, align=c]{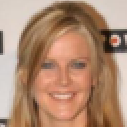} & 
\includegraphics[scale=0.3, align=c]{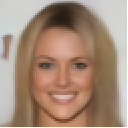} & 
\includegraphics[scale=0.3, align=c]{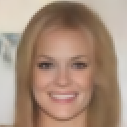} & 
\includegraphics[scale=0.3, align=c]{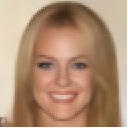} & 
\includegraphics[scale=0.3, align=c]{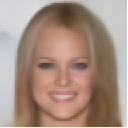} & 
\includegraphics[scale=0.3, align=c]{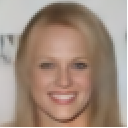} & 
\includegraphics[scale=0.3, align=c]{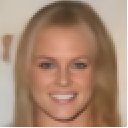} \\
\includegraphics[scale=0.3, align=c]{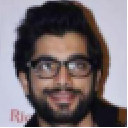} & 
\includegraphics[scale=0.3, align=c]{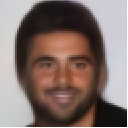} & 
\includegraphics[scale=0.3, align=c]{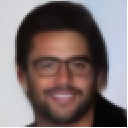} & 
\includegraphics[scale=0.3, align=c]{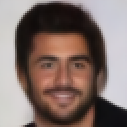} & 
\includegraphics[scale=0.3, align=c]{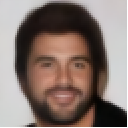} & 
\includegraphics[scale=0.3, align=c]{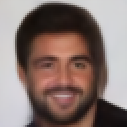} & 
\includegraphics[scale=0.3, align=c]{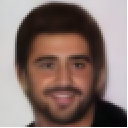} \\
\includegraphics[scale=0.3, align=c]{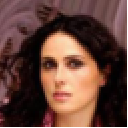} & 
\includegraphics[scale=0.3, align=c]{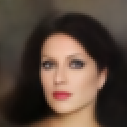} & 
\includegraphics[scale=0.3, align=c]{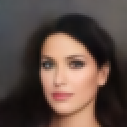} & 
\includegraphics[scale=0.3, align=c]{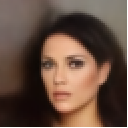} & 
\includegraphics[scale=0.3, align=c]{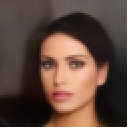} & 
\includegraphics[scale=0.3, align=c]{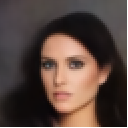} & 
\includegraphics[scale=0.3, align=c]{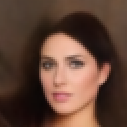} \\
\includegraphics[scale=0.3, align=c]{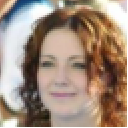} & 
\includegraphics[scale=0.3, align=c]{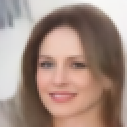} & 
\includegraphics[scale=0.3, align=c]{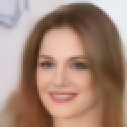} & 
\includegraphics[scale=0.3, align=c]{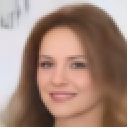} & 
\includegraphics[scale=0.3, align=c]{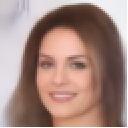} & 
\includegraphics[scale=0.3, align=c]{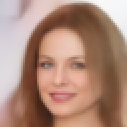} & 
\includegraphics[scale=0.3, align=c]{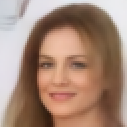} \\
\includegraphics[scale=0.3, align=c]{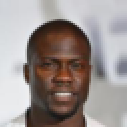} & 
\includegraphics[scale=0.3, align=c]{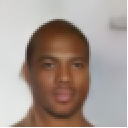} & 
\includegraphics[scale=0.3, align=c]{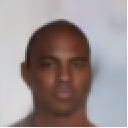} & 
\includegraphics[scale=0.3, align=c]{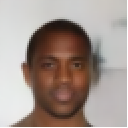} & 
\includegraphics[scale=0.3, align=c]{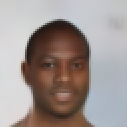} & 
\includegraphics[scale=0.3, align=c]{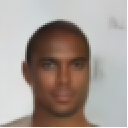} & 
\includegraphics[scale=0.3, align=c]{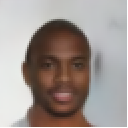} \\
\bottomrule
\end{tabular}

\end{table*}
\endgroup

\begingroup
\setlength{\tabcolsep}{1pt} %
\centering
\begin{table*}[htbp!]
\vspace{-1em}
\caption{CelebA 128x128 reconstructions at different compression rates using HQA, with number of transmitted bits.}
\vspace{0.5em}
\small

\centering
\begin{tabular}{cc|ccc}

\begin{tabular}{@{}c@{}}Original \\ 393,216 \end{tabular} & \quad &
\includegraphics[scale=0.6, align=c]{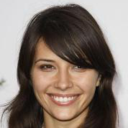} &
\includegraphics[scale=0.6, align=c]{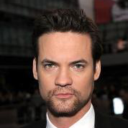} & 
\includegraphics[scale=0.6, align=c]{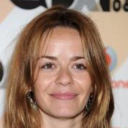} \\
\begin{tabular}{@{}c@{}}11x \\ 36,864 \end{tabular} & \quad &
\includegraphics[scale=0.6, align=c]{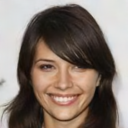} & 
\includegraphics[scale=0.6, align=c]{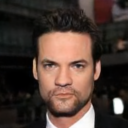} & 
\includegraphics[scale=0.6, align=c]{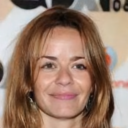} \\
\begin{tabular}{@{}c@{}}43x \\ 9,216 \end{tabular} & \quad &
\includegraphics[scale=0.6, align=c]{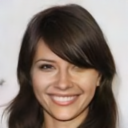} & 
\includegraphics[scale=0.6, align=c]{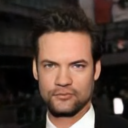} & 
\includegraphics[scale=0.6, align=c]{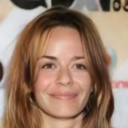} \\
\begin{tabular}{@{}c@{}}171x \\ 2,304 \end{tabular} & \quad &
\includegraphics[scale=0.6, align=c]{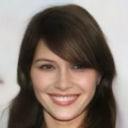} & 
\includegraphics[scale=0.6, align=c]{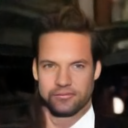} & 
\includegraphics[scale=0.6, align=c]{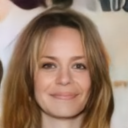} \\
\begin{tabular}{@{}c@{}}683x \\ 576 \end{tabular} & \quad &
\includegraphics[scale=0.6, align=c]{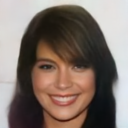} & 
\includegraphics[scale=0.6, align=c]{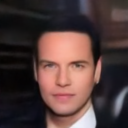} & 
\includegraphics[scale=0.6, align=c]{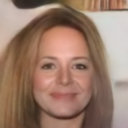} \\
\begin{tabular}{@{}c@{}}2,731x \\ 144 \end{tabular} & \quad &
\includegraphics[scale=0.6, align=c]{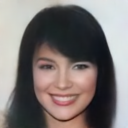} & 
\includegraphics[scale=0.6, align=c]{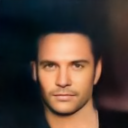} & 
\includegraphics[scale=0.6, align=c]{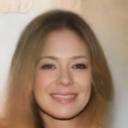} \\
\begin{tabular}{@{}c@{}}10,923x \\ 36 \end{tabular} & \quad &
\includegraphics[scale=0.6, align=c]{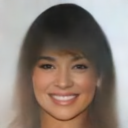} & 
\includegraphics[scale=0.6, align=c]{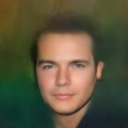} & 
\includegraphics[scale=0.6, align=c]{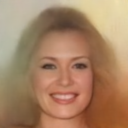} \\
\begin{tabular}{@{}c@{}}43,690x \\ 9 \end{tabular} & \quad &
\includegraphics[scale=0.6, align=c]{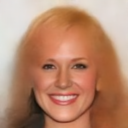} & 
\includegraphics[scale=0.6, align=c]{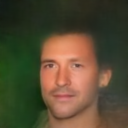} & 
\includegraphics[scale=0.6, align=c]{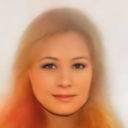} \\

\end{tabular}

\end{table*}
\endgroup

\begin{figure}[h!]
\centering
\makebox[\textwidth][c]{\includegraphics[scale=0.49]{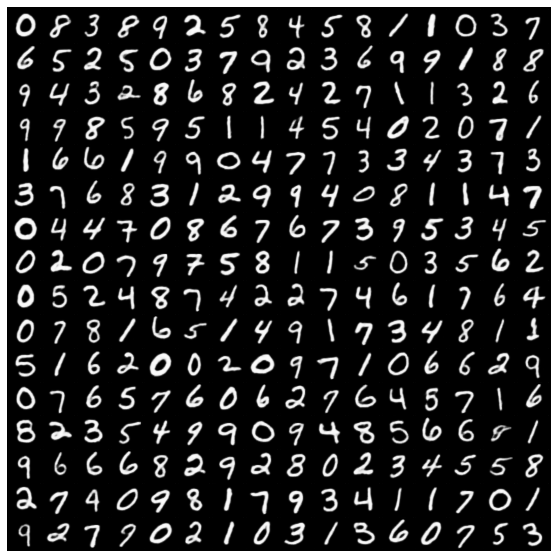}}
\caption{`Free' samples obtained by exhaustively enumerating over all 256 codes from the 1x1 latent space of the trained MNIST HQA stack and decoding into pixel-space.}
\end{figure}

\begin{figure}[h!]
\centering
\makebox[\textwidth][c]{\includegraphics[scale=0.49]{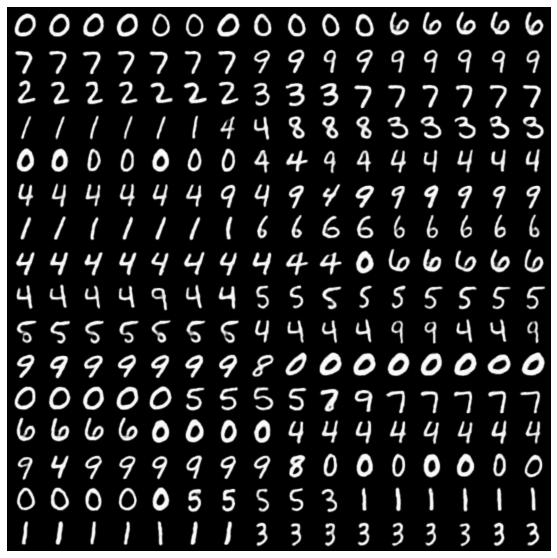}}
\caption{Rows show pairs of test images that have been encoded to the top of the HQA MNIST stack, interpolated across their codebook embeddings, quantized and then decoded.}
\end{figure}

\begin{figure}[h!]
\centering
\makebox[\textwidth][c]{\includegraphics[scale=0.48]{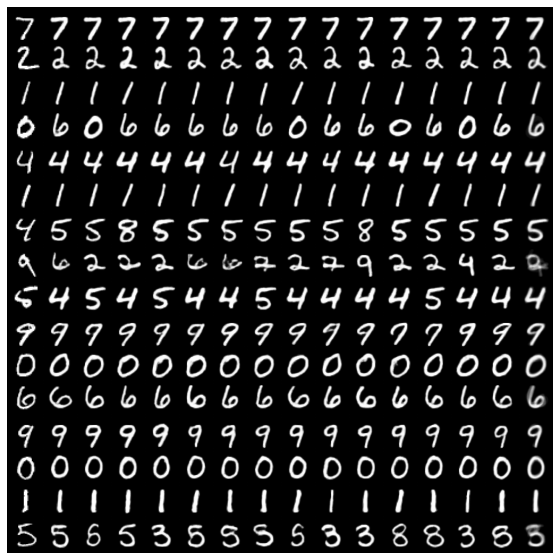}}
\caption{Each row displays the diversity of stochastic decoding for a different held out MNIST image. First column is the original, then 14 stochastic decodes, and then final column is 14 averaged decodes. Class switching behaviour is displayed due to the high compression factor with a 1x1 latent bottleneck.}
\end{figure}

\begin{figure}[h!]
\centering
\makebox[\textwidth][c]{\includegraphics[scale=0.48]{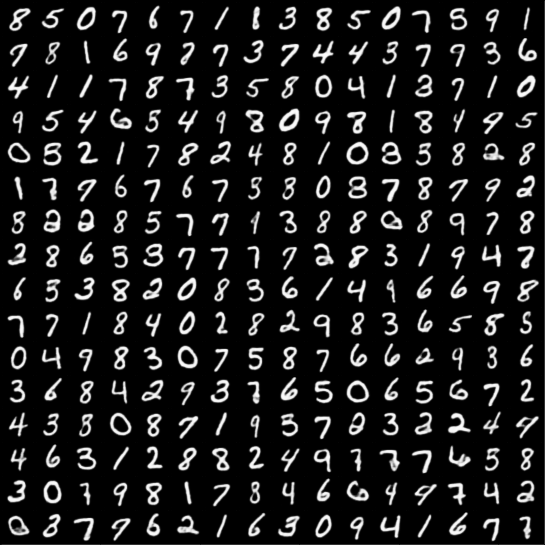}}
\caption{Samples generated by training a vanilla VAE on top of the learnt HQA 2x2 latent space and decoding first through the VAE then the HQA stack.}
\label{fig:mnistvae}
\end{figure}

\begin{table}[htbp]
\vspace{-0.5em}
\centering
\small
\tabcolsep=0.11cm
\caption{Interpolations generated for each layer in HQA. The far left and right images are originals. Others are decoded from the interpolated encoder output $z_e$. Bottom row (HQA-1) has a compression ratio of 4, each subsequent layer compresses by 4 again until the final layer (HQA-5) results in an 8 bit 1x1 latent space. Lower layers exhibit blurriness and overlapping versions of originals but higher layers have increasingly dense support allowing realistic and coherent looking digits from anywhere in the latent space.}
\label{tab:interp-layers}
\vspace{0.5em}
\begin{tabular}{m{1.2cm}|m{0.6cm}|m{4.7cm}|m{0.6cm}}
 \toprule
 System	&	Orig		&	Interpolation		& 	Orig	\\ \midrule
HQA-5	&	 \includegraphics[height=0.5cm,keepaspectratio]{mnist_results/interpolations/2-orig.png} 	&   \includegraphics[height=0.5cm,keepaspectratio]{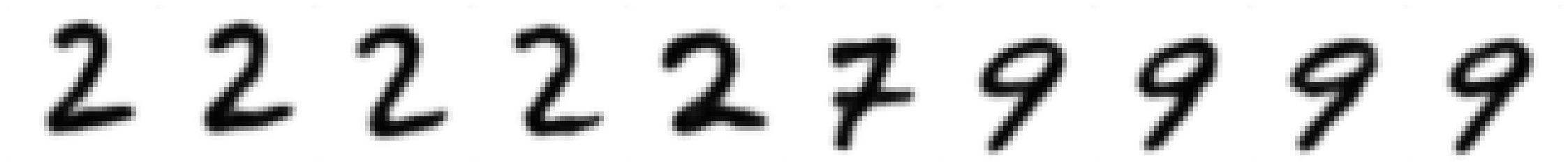}  &  \includegraphics[height=0.5cm,keepaspectratio]{mnist_results/interpolations/9-orig.png} \\ 
HQA-4	&	 \includegraphics[height=0.5cm,keepaspectratio]{mnist_results/interpolations/2-orig.png} 	&   \includegraphics[height=0.5cm,keepaspectratio]{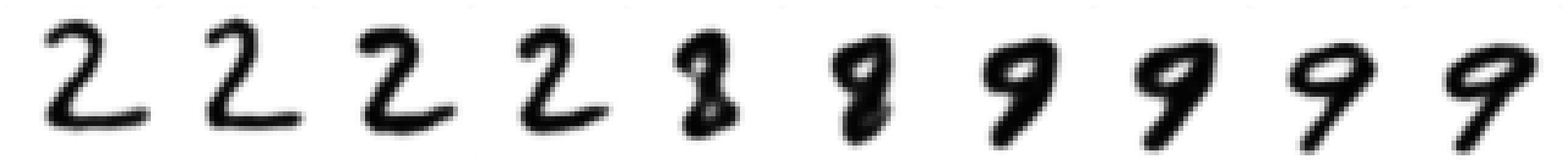}  &  \includegraphics[height=0.5cm,keepaspectratio]{mnist_results/interpolations/9-orig.png} \\ 
HQA-3	&	 \includegraphics[height=0.5cm,keepaspectratio]{mnist_results/interpolations/2-orig.png} 	&   \includegraphics[height=0.5cm,keepaspectratio]{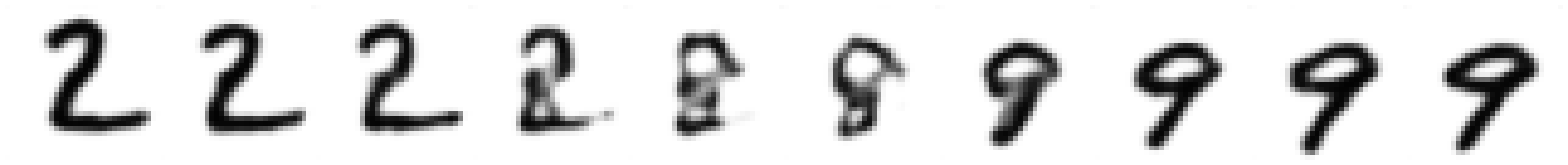}  &  \includegraphics[height=0.5cm,keepaspectratio]{mnist_results/interpolations/9-orig.png} \\ 
HQA-2	&	 \includegraphics[height=0.5cm,keepaspectratio]{mnist_results/interpolations/2-orig.png} 	&   \includegraphics[height=0.5cm,keepaspectratio]{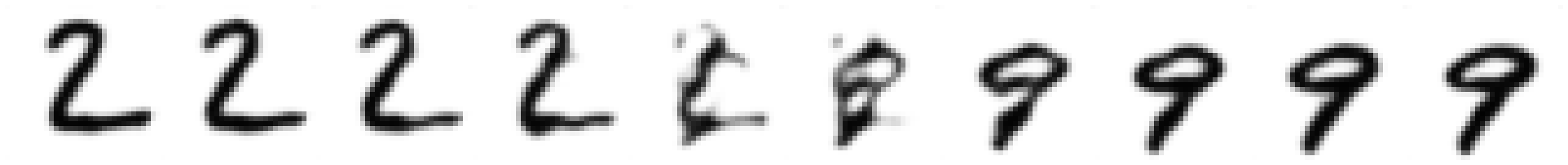}  &  \includegraphics[height=0.5cm,keepaspectratio]{mnist_results/interpolations/9-orig.png} \\ 
HQA-1	&	 \includegraphics[height=0.5cm,keepaspectratio]{mnist_results/interpolations/2-orig.png} 	&   \includegraphics[height=0.5cm,keepaspectratio]{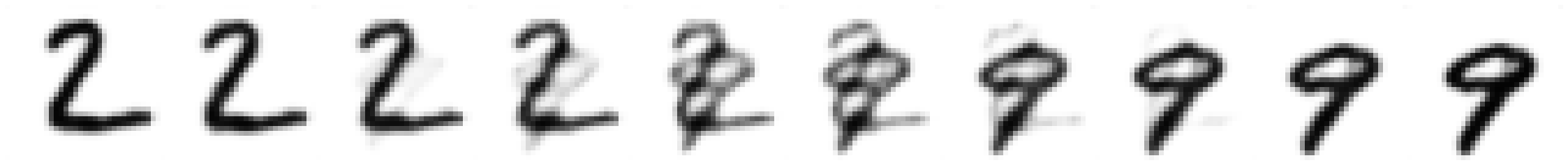}  &  \includegraphics[height=0.5cm,keepaspectratio]{mnist_results/interpolations/9-orig.png} \\ 
\end{tabular}
\vspace{-0.5em}
\end{table}

\section{Probabilistic VQ-VAE}
\label{sec:prop_vqvae}
\subsection{Motivation}
\noindent In this section we outline the probabilistic model that motivates the HQA loss:
\begin{equation}
    \label{eqn:our_loss_appendix}
    \mathcal{L} = - \log p(x|z=k) - \mathcal{H}[q(z|x)] + \mathbb{E}_{q(z|x)}||z_e(x) - e_z||_2^2 \: .
\end{equation}
A desired property of the HQA, motivated in Section~\ref{sec:stoch_motivation}, is the non-deterministic posterior $q(z|x)$ defined over codebook space. For the HQA, this is defined as a softmax with logits equal to the negative squared Euclidean distances between the encoded points ($z_e(x)$) and codebook vectors ($e_k$):
\begin{equation}
    \label{eqn:posterior}
    q(z=k|x) \propto \exp-||z_e(x) - e_k||_2^2 \: .
\end{equation}
 This form of posterior occurs in a simple Gaussian Mixture Model (GMM), where they are referred to as \emph{responsibilities}. In the GMM, the observed variables $x'$ are generated from possible sources $z' = {1,\dots,N}$. The responsibility of each source is then:
\begin{equation}
    \label{eqn:responsibilities}
    q(z'=k|x') \propto \exp-||x' - e_k||_2^2 \: .
\end{equation}
This mirrors Equation~\ref{eqn:posterior} where the encoded point $z_e(x)$ is replaced by the observations $x'$. Therefore, in order to derive a Evidence LOwer Bound (ELBO) for our model, we use a small extension to the GMM that incorporates the encoder-decoder architecture.

\subsection{Probabilistic Model}
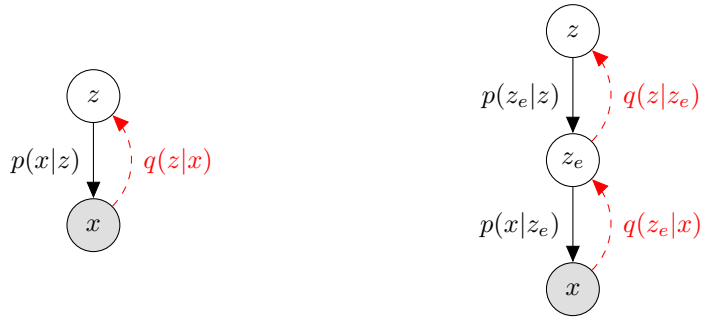
\begin{figure}[ht!]
    \centering
    \begin{subfigure}[t]{0.45\linewidth}
        \centering
        \raisebox{0.35\height}{\input{tikzdiagrams/ShallowMoG.tex}}
        \subcaption{Gaussian Mixture Model Network}
    \end{subfigure}
    \begin{subfigure}[t]{0.45\linewidth}
        \centering
        \input{tikzdiagrams/DeepMoG.tex}
        \subcaption{A single layer of the HQA as a Bayesian Network}
    \end{subfigure}
    \caption{Contrasting the probabilistic model of a GMM and a single layer of the HQA Inference distributions are shown in red. \\\hspace{\textwidth}}
    \label{fig:models}
\end{figure}
We introduce an additional latent variable $z_e$ into the standard GMM setup, so that the distribution $p(x|z)$ factorizes as: 
\begin{equation}
    \label{eqn:factorization}
    p(x|z) = \underbrace{p(x|z_e)}_{\text{Decoder}}\underbrace{p(z_e|z)}_{\text{GMM}} \:.
\end{equation}
We contrast these two models in Figure~\ref{fig:models}. In this setup we treat $z_e$ as being generated from a GMM. $z_e$ is then fed through the decoder neural network.  \\

\noindent To then infer a value for $z$ we first approximate the posterior $p(z_e|x)$ with a deterministic distribution on the output of the encoder neural network. To emphasize this in our analysis we refer to the output of the encoder as $z_e(x)$, whilst we refer to the latent variable as $z_e$. The final stage of inference to calculate $p(z|z_e)$ reduces to a simple GMM model with observed variables $x'$ in Equation~\ref{eqn:responsibilities} replaced with $z_e(x)$. This leads exactly to the posterior probabilities given in Equation~\ref{eqn:posterior}. As $q(z_e|x)$ is deterministic we have that $q(z|z_e) = q(z|x)$ and so we use these expressions interchangeably. \\

\noindent This model is a Variational Autoencoder with a simple Mixture of Gaussians prior. In the prior, each Gaussian is assumed to be independent and have constant variance. Similar, more complex models are considered in \citet{dilokthanakul2016deep, nalisnick2016approximate, VAEVamp}.

\subsection{Deriving the ELBO}
Finally, as we have recovered the posterior probabilities we desire, we now derive the ELBO loss. For a general latent variable model with observation $x$ this is formulated as:
\begin{equation}
    \mathcal{L}_{\text{ELBO}} = \mathbb{E}_{q(z|x)}\log p_\theta(x|z)  - \text{KL}[q(z|x) || p(z)] \:
\end{equation}
where $q(z|x)$ is our approximate posterior distribution. However, in our case we have two latent variables, giving the loss:
\begin{equation}
    \mathcal{L}_{\text{ELBO}} = \mathbb{E}_{q(z, z_e|x)}\log p_\theta(x|z,z_e) - \text{KL}[q(z, z_e|x) || p(z,z_e)] \: .
\end{equation}
We can then make use of the factorization in Equation~\ref{eqn:factorization} to rearrange this as:
\begin{equation}
    \mathcal{L}_{\text{ELBO}} = \underbrace{\mathbb{E}_{q(z|z_e)q(z_e|x)}\log p_\theta(x|z_e)}_{\text{Reconstruction Loss}} - \underbrace{\text{KL}[q(z| z_e)q(z_e|x) || p(z) p(z_e|z)]}_{\text{KL to prior}} \: .
\end{equation}
We now consider each of these terms separately.
\subsubsection{Prior KL Loss}
The Prior KL Loss is given by:
\begin{equation}
    \mathcal{L}_{\text{prior}} = \text{KL}[q(z| z_e)q(z_e|x) || p(z) p(z_e|z)] \:.
\end{equation}
This factorizes into two separate KL terms
\begin{equation}
    \mathcal{L}_{\text{prior}} = \text{KL}[q(z| x)|| p(z)] + \mathbb{E}_{q(z|x)}\text{KL}[q(z_e|x) || p(z_e|z)] \: .
\end{equation}
As we define a uniform prior over mixture parameters $p(z)$, the first term becomes the entropy term $\mathcal{H}(q(z|x))$ as given in Equation~\ref{eqn:our_loss_appendix}. The next term is then:
\begin{equation}
    \mathbb{E}_{q(z|x)}\text{KL}[q(z_e|x) || p(z_e|z)] = - \mathbb{E}_{q(z|x)}\log\left(e^{ -||z_e(x) - e_z||_2^2}\right) \\
    = \mathbb{E}_{q(z|x)}||z_e(x) - e_z||_2^2 \:
\end{equation}
which is the final part of Equation~\ref{eqn:our_loss_appendix}. We omit two details: the constant terms and the factor of $0.5$ multiplied by the variance that usually occurs in the Gaussian density function as this is reweighted before training. 
\subsubsection{Reconstruction Loss}
The reconstruction loss is given by:
\begin{equation}
    \mathcal{L}_{\text{recon}} = \mathbb{E}_{q(z|z_e)q(z_e|x)}\log p_\theta(x|z_e) = \mathbb{E}_{q(z_e|x)}\mathbb{E}_{q(z|z_e)}\log p_\theta(x|z_e) \: .
\end{equation}
In order to train with the quantized behaviour we require, we don't follow this calculation when calculating the reconstruction loss. Instead we sample from $q(z|z_e(x))$ and feed this back through the decoder. This modification gives
\begin{equation}
    \mathcal{L'}_{\text{recon}} = \log p(x|z_e = k) \:
\end{equation}
where $k$ is sampled from $q(z|z_e(x))$. To clarify, whilst training, instead of using the encoded point $z_e$ as the input to the decoder, we feed the codebook vector sampled from the posterior $q(z|x)$. 

\subsection{VQ-VAE as a limiting case}
If we include a temperature parameter in our softmax posterior
\begin{equation}
    \label{eqn:tempposterior}
    q(z=k|x) \propto \exp\left(-\frac{1}{\tau}||z_e(x) - e_k||_2^2\right) \:
\end{equation}
then as $\tau \rightarrow 0$, the posterior converges to a deterministic distribution:
\begin{equation}
    \label{eqn:deterministicpos}
    q(z = k|x) = \begin{cases}1 \quad\text{for} \enspace k = \text{argmin}_j ||z_e(x) - e_j||_2\\ 0\quad \text{otherwise}\end{cases}
\end{equation}
This is precisely the posterior that arises in the VQ-VAE. In addition, the KL prior terms then become:
\begin{equation}
    \label{eqn:limitentropy}
    \mathcal{H}(q(z|x)) = 0
\end{equation}
\begin{equation}
    \label{eqn:limitcommit}
    \mathbb{E}_{q(z|x)}||z_e(x) - e_z||_2^2 = ||z_e(x) - e_k||_2^2
\end{equation}
If then stop gradient operators are applied to \eqref{eqn:limitcommit}, the commitment and codebook loss from the VQ-VAE are recovered.

\section{Architecture, training and hyper-parameters}
\label{sec:arch_hparams}
\subsection{HQA}

Each layer in the HQA stack is composed of an encoder, decoder and vector quantization layer. Encoders and decoders are feed forward networks composed of convolutional layers with 3x3 filters. Optional dilated convolutions are used in the decoder to increase the decoder's receptive field. Each code in the VQ layer codebook is represented by a 64 dimensional vector. The input $\hat{z_e}$ to layers 2 and above are normalized using running statistics, which was shown to stabilise training. A $\operatorname{sigmoid}$ activation is applied to the output of the decoder in the first layer.

The downsampling needed for compression is achieved through a strided convolution in the encoder and upsampling through nearest neighbour interpolation in the decoder. Each HQA layer is trained greedily with an MSE loss; gradients are only back-propagated through that single layer. For the first layer, the loss is taken between input pixels and decoder outputs, while all other layers calculate the loss between the input embedding $z_e$ and the predicted $\hat{z_e}$. 

Optimization is performed using RAdam \cite{liu2019variance} with a learning rate of 4e-4 which is cosine annealed in the final third of training. Each layer was trained with distributed training across 8 Nvidia TITAN RTX's for CelebA, whilst MNIST was trained on a single TITAN X. During training, the Gumbel softmax temperature is linearly annealed to 0.01, with an initial temperature of 0.4 and 0.66 for CelebA and MNIST respectively.

\begin{table*}[!htbp]
\centering
\small
\caption{Hyper parameters of HQA network used for CelebA experiment}
\vspace{-1.0em}
\begin{tabular}{lccccccc}
\addlinespace
& L1 & L2 & L3 & L4 & L5 & L6 & L7  \\
\toprule
Input size & 64 & 64 & 32 & 16 & 8 & 4 & 2  \\
Batch size & 1024 & 1024 & 1024 & 1024 & 1024 & 1024 & 1024 \\
Encoder layers & 3 & 3 & 3 & 3 & 3 & 3 & 3  \\
Decoder layers & 6 & 6 & 6 & 6 & 6 & 6 & 6  \\
Encoder hidden units & 64 & 64 & 512 & 512 & 512 & 512 & 512 \\
Decoder hidden units & 64 & 64 & 512 & 512 & 512 & 512 & 512  \\
Codebook size & 512 & 512 & 512 & 512 & 512 & 512 & 512 \\
$\beta_e$ (entropy loss coefficient) & 5e-5 & 5e-5 & 5e-5 & 5e-5 & 5e-5 & 5e-5 & 5e-5 \\
$\beta_c$ (commitment loss coefficient) & 5e-5 & 5e-5 & 5e-5 & 5e-5 & 5e-5 & 5e-5 & 5e-5 \\
Training steps & 100k & 100k & 100k & 100k & 60k & 30k & 30k \\
Dropout & 0.0 & 0.0 & 0.0 & 0.5 & 0.5 & 0.5 & 0.5 \\

\end{tabular}
\end{table*}

\begin{table*}[!htbp]
\centering
\small
\caption{Hyper parameters of HQA network used for MNIST experiment}
\vspace{-1.0em}
\begin{tabular}{lccccc}
\addlinespace
&  L1 & L2 & L3 & L4 & L5 \\
\toprule
Input size & 32 & 16 & 8 & 4 & 2 \\
Batch size & 512 & 512 & 512 & 512 & 512 \\
Encoder layers & 3 & 3 & 3 & 3 & 3 \\
Decoder layers & 3 & 3 & 3 & 3 & 3  \\
Encoder hidden units & 16 & 16 & 32 & 48 & 80 \\
Decoder hidden units & 16 & 32 & 48 & 80 & 128 \\
Codebook size & 256 & 256 & 256 & 256 & 256 \\
$\beta_e$ (entropy loss coefficient) & 1e-3 & 1e-3 & 1e-3 & 1e-3 & 1e-3 \\
$\beta_c$ (commitment loss coefficient) & 1e-3 & 1e-3 & 1e-3 & 1e-3 & 1e-3 \\
Training steps & 18k & 18k & 18k & 18k & 18k \\

\end{tabular}
\end{table*}

\subsection{HAMs}

The implemented HAMs architecture follows \citet{HAMS}. Notably, it implements an MSE loss on pixels but all other layers use cross entropy for the reconstruction term. Separate commitment and codebook loss terms are also used. The codebook is not learnt directly, but updated via an online exponential moving average version of k-means. For the CelebA experiment a smaller batch sizes where used than the 1024 used for HQA. This is because we found training of HAMs to be very unstable if large batch sizes were used.

\begin{table*}[!htbp]
\centering
\small
\caption{Hyper parameters of HAMs network used for CelebA experiment}
\vspace{-1.0em}
\begin{tabular}{lccccccc}
\addlinespace

& L1 & L2 & L3 & L4 & L5 & L6 & L7 \\
\toprule
Input size & 64 & 64 & 32 & 16 & 8 & 4 & 2\\
Batch size & 32 & 64 & 64 & 64 & 64 & 64 & 64  \\
Encoder conv layers & 3 & 3 & 3 & 3 & 3 & 3 & 3  \\
Decoder conv layers & 3 & 3 & 3 & 3 & 3 & 3 & 3  \\
Encoder hidden units & 64 & 80 & 256 & 256 & 256 & 256 & 512\\
Decoder hidden units & 64 & 80 & 512 & 512 & 512 & 512 & 512 \\
Encoder residual blocks & 2 & 2 & 2 & 3 & 3 & 2 & 1 \\
Decoder residual blocks & 2 & 2 & 2 & 3 & 3 & 2 & 1   \\
Codebook size & 512 & 512 & 512 & 512 & 512 & 512 & 512 \\
$\beta$ (commitment loss coefficient) & 1 & 50 & 50 & 50 & 50 & 50 & 10  \\
Learning rate & 4e-4 & 4e-4 & 4e-4 & 4e-4 & 1e-4 & 1e-4 & 1e-4\\
Training steps & 250k & 300k & 50k & 50k & 50k & 50k & 25k \\
\end{tabular}
\end{table*}

\begin{table*}[!htbp]
\centering
\small
\caption{Hyper parameters of HAMs network used for MNIST experiment}
\vspace{-1.0em}
\begin{tabular}{lccccc}
\addlinespace

& L1 & L2 & L3 & L4 & L5  \\
\toprule
Input size & 32 & 16 & 8 & 4 & 2 \\
Batch size & 256 & 256 & 256 & 256 & 256  \\
Encoder conv layers & 3 & 3 & 3 & 3 & 3   \\
Decoder conv layers & 3 & 3 & 3 & 3 & 3  \\
Encoder hidden units & 16 & 16 & 32 & 48 & 80 \\
Decoder hidden units & 16 & 26 & 40 & 58 & 96 \\
Encoder residual blocks & 0 & 0 & 0 & 0 & 0 \\
Decoder residual blocks & 0 & 0 & 0 & 0 & 0  \\
Codebook size & 256 & 256 & 256 & 256 & 256 \\
$\beta$ (commitment loss coefficient) & 0.02 & 0.02 & 0.02 & 0.02 & 0.02 \\
Learning rate & 4e-4 & 4e-4 & 4e-4 & 4e-4 & 1e-4 \\
Training steps &  18k & 18k & 18k & 18k & 18k \\
\end{tabular}
\vspace{-1em}
\end{table*}

\subsection{VQ-VAE}

The implemented VQ-VAE \cite{VQVAE} architecture is comparable to HAMs, with the noticeable exception that there is no hierarchy. The same compression rates are achieved through downsampling multiple times. The entire network is trained end-to-end as a single layer, instead of greedily with local losses. The layers denoted in the table below refer VQ-VAE systems with equivalent compression factors to the same HQA and HAM layers. In all instances predictions are made in pixel space. The residual block implementation is based on the original VQ-VAE. As with HAMs, small batch sizes had to be used for the CelebA experiment as large batch sizes lead to instability.

\begin{table*}[!htbp]
\centering
\small
\caption{Hyper parameters of VQ-VAE network used for CelebA experiment}
\vspace{-1.0em}
\begin{tabular}{lccccccc}
\addlinespace
& L1 & L2 & L3 & L4 & L5 & L6 & L7 \\
\toprule
Input size & 64 & 64 & 32 & 16 & 8 & 4 & 2  \\
Batch size & 32 & 64 & 64 & 64 & 64 & 64 & 64  \\
Encoder conv layers & 2 & 3 & 4 & 5 & 6 & 7 & 8 \\
Decoder conv layers & 3 & 4 & 5 & 6 & 7 & 8 & 9  \\
Encoder hidden units & 64 & 80 & 256 & 256 & 384 & 400 & 512 \\
Decoder hidden units & 64 & 80 & 256 & 512 & 512 & 512 & 512\\
Encoder residual blocks & 2 & 3 & 4 & 4 & 4 & 4 & 2 \\
Decoder residual blocks & 2 & 3 & 4 & 4 & 4 & 4 & 2  \\
Codebook size & 512 & 512 & 512 & 512 & 512 & 512 & 512 \\
$\beta$ (commitment loss coefficient) & 0.05 & 0.25 & 0.25 & 0.25 & 0.25 & 0.25 & 0.25  \\
Learning rate & 4e-5 & 4e-5 & 4e-5 & 1e-4 & 1e-4 & 1e-4 & 1e-4 \\
Training steps & 250k & 250k & 250k & 150k & 150k & 150k & 50k  \\
\end{tabular}
\end{table*}

\begin{table*}[!htbp]
\centering
\small
\caption{Hyper parameters of VQ-VAE network used for MNIST experiment}
\vspace{-1.0em}
\begin{tabular}{lcccccccccccc}
\addlinespace
& L1 & L2 & L3 & L4 & L5 \\
\toprule
Input size  & 32 & 16 & 8 & 4 & 2 \\
Batch size & 512 & 512 & 512 & 512 & 512 \\
Encoder conv layers &  2 & 3 & 4 & 5 & 6 \\
Decoder conv layers & 3 & 4 & 5 & 6 & 7  \\
Encoder hidden units & 22 & 40 & 50 & 62 & 78 \\
Decoder hidden units  & 16 & 18 & 20 & 22 & 22 \\
Encoder residual blocks & 0 & 0 & 0 & 0 & 0 \\
Decoder residual blocks & 0 & 0 & 0 & 0 & 0  \\
Codebook size &  256 & 256 & 256 & 256 & 256 \\
$\beta$ (commitment loss coefficient) &  0.125 & 0.125 & 0.125 & 0.125 & 0.125 \\
Learning rate & 4e-4 & 4e-4 & 4e-4 & 4e-4 & 4e-4 \\
Training steps & 18k & 18k & 18k & 18k & 18k \\
\end{tabular}
\end{table*}

\subsection{Codebook Resetting}
During training, the total number of times that $z_e$ is quantized to each code is accumulated over 20 batches. After these 20 batches, the most and least used code, $e_m$ and $e_l$ respectively, are found. If the usage of $e_l$ is less than 3\% than that of $e_m$, the position of $e_l$ is reset such that $e_l := e_m + \epsilon$ where $\epsilon \sim N(0, 0.01)$. This scheme is activate for the first 75\% of training.

\clearpage

\section{Algorithm description}
\label{sec:algorithm}

\begin{algorithm}[h!]
\caption{HQA Training}
\label{alg:training}
\begin{algorithmic}[1]
\State $\bm{e}$: codebook embeddings, $e_k$: embdding for code $k$, $N$: number of codes in each layer
\State $L$: number of layers in stack
\State \(\theta_i \gets\) Initialize network parameters for encoders ($Encoder_{i}$) and decoders ($Decoder_{i}$) $\forall i \in L$

\For{$l$ in $1,\ldots,L$} \Comment{Train each layer greedily}
\State $\tau \gets 0.4$ \Comment{Set initial codebook temperature}
\While{not converged}
\State \(X \gets\) Random minibatch

\If{$l = 1$}
    \State \(\bm{z_{e-lower}} \gets\) X
\Else
    \State \(\bm{z_{e-lower}} \gets\) $Encoder_{0..l-1}(X)$ \Comment{Encode up through pre-trained lower layers - no quantization}
\EndIf

\State \(\bm{z_e} \gets\) $Encoder_{l}(\bm{z_{e-lower}})$
\State $p(k|\bm{z_e}) = \exp\left(-\frac{1}{2}||\bm{z_e} - \bm{e}_k||_2^2\right) / \sum_{i=1}^N \exp\left(-\frac{1}{2}||\bm{z_e} - \bm{e}_i||_2^2\right))$ \Comment{Distribution over codes}
\State \(\text{softonehot} \sim \) $\text{RelaxedCategorical}(\tau, p(k|\bm{z_e}))$ \Comment{Reparameterized Gumbel-softmax sample}
\State \(\bm{z_{q-soft}} \gets\) $\text{softonehot} * \bm{e}$ \Comment{Soft quantized codebook lookup}
\State \( \bm{\hat{z}_{e-lower}} \gets\) $Decoder_{l}(\bm{z_{q-soft}})$

\State $\mathcal{L'}_{recon} = \left(\bm{\hat{z}_{e-lower}} - \bm{z_{e-lower}}\right)^2$
\State $\mathcal{L}_{entropy} = \sum_k p(k|\bm{z_e}) \log p(k|\bm{z_e})$
\State $\mathcal{L}_{commit} = \sum_k p(k|\bm{z_e})  ||\bm{z_e} - e_k||_2^2$

\State $\theta_i \gets \theta_i - \eta\nabla_{\theta_i} (\mathcal{L'}_{recon} +  \beta_e \mathcal{L}_{entropy} + \beta_c \mathcal{L}_{commit} ))$
\State \(\tau \gets anneal(\tau) \) \Comment{Anneal linearly}
\EndWhile
\EndFor
\end{algorithmic}
\end{algorithm}

\begin{algorithm}[h!]
\caption{HQA Reconstruction}
\label{alg:reconstruct}
\begin{algorithmic}[1]
\State $\bm{e}$: codebook embeddings
\State $L$: number of layers in stack
\State Trained encoders ($Encoder_{i}$) and decoders ($Decoder_{i}$) $\forall i \in L$
\State $x$: Datapoint to reconstruct
\State \(\bm{z_e} \gets\) $Encoder_{0..l}(x)$

\For{$l$ in $L,\ldots,1$}
\State $p(k|\bm{z_e}) = \exp\left(-\frac{1}{2}||\bm{z_e} - \bm{e}_k||_2^2\right) / \sum_{i=1}^N \exp\left(-\frac{1}{2}||\bm{z_e} - \bm{e}_i||_2^2\right))$ \Comment{Distribution over codes}
\State \(\text{onehot} \sim \) $ p(k|\bm{z_e})$
\State \(\bm{z_q} \gets\) $\text{onehot}* \bm{e}$ \Comment{Hard-quantized codebook lookup}
\State \(\bm{z_e} \gets Decoder_{l}(\bm{z_q}) \)
\EndFor
\State \Return $\bm{z_e}$
\end{algorithmic}
\end{algorithm}

Note that for hard reconstructions at fixed rates, we do not necessarily need to perform hard-quantized codebook lookups except on the very top codebook. For simplicity, and to provide a single hierarchy where each layer can provide compression at a fixed rate, we anneal the temperature close to zero and at test time always perform hard quantization operations at each layer as outlined in Algorithm~\ref{alg:reconstruct}.

\end{document}

%% file: tikzdiagrams/toy_data_graph_1.tex
\begin{tikzpicture}[xscale=0.46,yscale=0.46,every node/.append style={font=\Large}]
        \begin{axis}[
            ylabel near ticks,
            xlabel={\huge$x$},
            ylabel={Density},
            xmin=-0.8, xmax=0.8,
            ymin=-0.2, ymax=11,
            xticklabel style={/pgf/number format/fixed},
            ticks=none,
            legend pos=north east,
            ymajorgrids=false,
        ]
        \addplot[,color=mycolour1,mark size=4pt,]
            coordinates {
               (-0.8, 0.0)(-0.7946127946127947, 0.0)(-0.7892255892255893, 0.0)(-0.7838383838383839, 0.0)(-0.7784511784511785, 0.0)(-0.7730639730639731, 0.0)(-0.7676767676767677, 0.0)(-0.7622895622895624, 0.0)(-0.756902356902357, 0.0)(-0.7515151515151516, 0.0)(-0.7461279461279462, 0.0)(-0.7407407407407408, 0.0)(-0.7353535353535354, 0.0)(-0.72996632996633, 0.0)(-0.7245791245791247, 0.0)(-0.7191919191919193, 0.0)(-0.7138047138047139, 0.0)(-0.7084175084175085, 0.0)(-0.7030303030303031, 0.0011601562499999997)(-0.6976430976430976, 0.0)(-0.6922558922558923, 0.0)(-0.6868686868686869, 0.0)(-0.6814814814814816, 0.003480468749999999)(-0.6760942760942761, 0.010441406249999998)(-0.6707070707070708, 0.015082031249999997)(-0.6653198653198653, 0.041765624999999994)(-0.65993265993266, 0.08701171874999998)(-0.6545454545454545, 0.15894140624999997)(-0.6491582491582492, 0.33412499999999995)(-0.6437710437710438, 0.6380859374999999)(-0.6383838383838384, 1.0186171874999999)(-0.632996632996633, 1.4815195312499998)(-0.6276094276094276, 2.3075507812499993)(-0.6222222222222222, 3.1208203124999994)(-0.6168350168350168, 3.859839843749999)(-0.6114478114478115, 4.475882812499999)(-0.6060606060606061, 4.868015624999999)(-0.6006734006734007, 5.022316406249999)(-0.5952861952861953, 4.630183593749999)(-0.5898989898989899, 4.038503906249999)(-0.5845117845117845, 3.2832421874999995)(-0.5791245791245792, 2.5175390624999996)(-0.5737373737373738, 1.7205117187499996)(-0.5683501683501684, 1.1996015624999998)(-0.562962962962963, 0.6868124999999998)(-0.5575757575757576, 0.41997656249999993)(-0.5521885521885522, 0.20998828124999996)(-0.5468013468013468, 0.11253515624999998)(-0.5414141414141415, 0.04640624999999999)(-0.5360269360269361, 0.017402343749999997)(-0.5306397306397307, 0.011601562499999997)(-0.5252525252525253, 0.0011601562499999997)(-0.5198653198653199, 0.0)(-0.5144781144781145, 0.0)(-0.509090909090909, 0.0)(-0.5037037037037038, 0.0)(-0.4983164983164983, 0.0)(-0.49292929292929294, 0.0011601562499999997)(-0.48754208754208755, 0.0011601562499999997)(-0.48215488215488217, 0.0023203124999999995)(-0.4767676767676768, 0.0011601562499999997)(-0.4713804713804714, 0.011601562499999997)(-0.465993265993266, 0.026683593749999995)(-0.46060606060606063, 0.07076953124999999)(-0.45521885521885525, 0.16242187499999997)(-0.44983164983164986, 0.27611718749999997)(-0.4444444444444445, 0.5394726562499998)(-0.4390572390572391, 1.0186171874999999)(-0.4336700336700337, 1.5070429687499998)(-0.4282828282828283, 2.2274999999999996)(-0.42289562289562294, 3.0778945312499992)(-0.41750841750841755, 3.6753749999999994)(-0.41212121212121217, 4.381910156249999)(-0.4067340067340068, 4.764761718749999)(-0.4013468013468014, 5.060601562499999)(-0.39595959595959596, 4.692832031249999)(-0.39057239057239057, 4.202085937499999)(-0.3851851851851852, 3.3064453124999993)(-0.3797979797979798, 2.6115117187499997)(-0.3744107744107744, 1.8040429687499997)(-0.36902356902356903, 1.2750117187499999)(-0.36363636363636365, 0.7366992187499999)(-0.35824915824915826, 0.4791445312499999)(-0.3528619528619529, 0.24479296874999995)(-0.3474747474747475, 0.09513281249999998)(-0.3420875420875421, 0.06496874999999999)(-0.3367003367003367, 0.026683593749999995)(-0.33131313131313134, 0.010441406249999998)(-0.32592592592592595, 0.003480468749999999)(-0.32053872053872057, 0.003480468749999999)(-0.3151515151515152, 0.0011601562499999997)(-0.3097643097643098, 0.0)(-0.3043771043771044, 0.0)(-0.29898989898989903, 0.0)(-0.29360269360269364, 0.0)(-0.28821548821548826, 0.0)(-0.2828282828282829, 0.0)(-0.2774410774410775, 0.0)(-0.2720538720538721, 0.0)(-0.2666666666666667, 0.0)(-0.26127946127946133, 0.0)(-0.25589225589225595, 0.0)(-0.25050505050505056, 0.0)(-0.24511784511784518, 0.0)(-0.2397306397306398, 0.0)(-0.2343434343434344, 0.0)(-0.22895622895622902, 0.0)(-0.22356902356902353, 0.0)(-0.21818181818181814, 0.0)(-0.21279461279461276, 0.0)(-0.20740740740740737, 0.0)(-0.202020202020202, 0.0)(-0.1966329966329966, 0.0)(-0.19124579124579122, 0.0)(-0.18585858585858583, 0.0)(-0.18047138047138045, 0.0)(-0.17508417508417506, 0.0)(-0.16969696969696968, 0.0)(-0.1643097643097643, 0.0)(-0.1589225589225589, 0.0)(-0.15353535353535352, 0.0)(-0.14814814814814814, 0.0)(-0.14276094276094276, 0.0)(-0.13737373737373737, 0.0)(-0.13198653198653199, 0.0)(-0.1265993265993266, 0.0)(-0.12121212121212122, 0.0)(-0.11582491582491583, 0.0)(-0.11043771043771045, 0.0)(-0.10505050505050506, 0.0)(-0.09966329966329968, 0.0)(-0.09427609427609429, 0.0)(-0.0888888888888889, 0.0)(-0.08350168350168352, 0.0)(-0.07811447811447814, 0.0)(-0.07272727272727275, 0.0)(-0.06734006734006737, 0.0)(-0.06195286195286198, 0.0)(-0.0565656565656566, 0.0)(-0.05117845117845121, 0.0)(-0.04579124579124583, 0.0)(-0.04040404040404044, 0.0)(-0.03501683501683506, 0.0)(-0.029629629629629672, 0.0)(-0.024242424242424288, 0.0)(-0.018855218855218903, 0.0)(-0.013468013468013518, 0.0)(-0.008080808080808133, 0.0)(-0.002693602693602748, 0.0)(0.002693602693602637, 0.0)(0.008080808080808133, 0.0)(0.013468013468013518, 0.0)(0.018855218855218903, 0.0)(0.024242424242424288, 0.0)(0.029629629629629672, 0.0)(0.03501683501683506, 0.0)(0.04040404040404044, 0.0)(0.04579124579124583, 0.0)(0.05117845117845121, 0.0)(0.0565656565656566, 0.0)(0.06195286195286198, 0.0)(0.06734006734006737, 0.0)(0.07272727272727275, 0.0)(0.07811447811447814, 0.0)(0.08350168350168352, 0.0)(0.0888888888888889, 0.0)(0.09427609427609429, 0.0)(0.09966329966329968, 0.0)(0.10505050505050506, 0.0)(0.11043771043771045, 0.0)(0.11582491582491583, 0.0)(0.12121212121212122, 0.0)(0.1265993265993266, 0.0)(0.13198653198653199, 0.0)(0.13737373737373737, 0.0)(0.14276094276094276, 0.0)(0.14814814814814814, 0.0)(0.15353535353535352, 0.0)(0.1589225589225589, 0.0)(0.1643097643097643, 0.0)(0.16969696969696968, 0.0)(0.17508417508417506, 0.0)(0.18047138047138045, 0.0)(0.18585858585858583, 0.0)(0.19124579124579122, 0.0)(0.1966329966329966, 0.0)(0.202020202020202, 0.0)(0.20740740740740748, 0.0)(0.21279461279461276, 0.0)(0.21818181818181825, 0.0)(0.22356902356902353, 0.0)(0.22895622895622902, 0.0)(0.2343434343434343, 0.0)(0.2397306397306398, 0.0)(0.24511784511784507, 0.0)(0.25050505050505056, 0.0)(0.25589225589225584, 0.0)(0.26127946127946133, 0.0)(0.2666666666666666, 0.0)(0.2720538720538721, 0.0)(0.2774410774410774, 0.0)(0.2828282828282829, 0.0)(0.28821548821548815, 0.0)(0.29360269360269364, 0.0)(0.2989898989898989, 0.0)(0.3043771043771044, 0.0)(0.3097643097643097, 0.0)(0.3151515151515152, 0.0)(0.32053872053872046, 0.0011601562499999997)(0.32592592592592595, 0.008121093749999999)(0.3313131313131312, 0.029003906249999996)(0.3367003367003367, 0.038285156249999994)(0.342087542087542, 0.12993749999999998)(0.3474747474747475, 0.22391015624999996)(0.352861952861953, 0.4211367187499999)(0.35824915824915826, 0.7645429687499998)(0.36363636363636376, 1.2100429687499998)(0.36902356902356903, 1.8875742187499995)(0.3744107744107745, 2.7530507812499994)(0.3797979797979798, 3.372574218749999)(0.3851851851851853, 4.182363281249999)(0.39057239057239057, 4.6243828124999995)(0.39595959595959607, 5.044359374999999)(0.40134680134680134, 4.803046874999999)(0.40673400673400684, 4.513007812499999)(0.4121212121212121, 3.842437499999999)(0.4175084175084176, 2.9943632812499996)(0.4228956228956229, 2.2228593749999996)(0.4282828282828284, 1.5105234374999996)(0.43367003367003365, 0.9652499999999998)(0.43905723905723915, 0.5905195312499999)(0.4444444444444444, 0.31904296874999993)(0.4498316498316499, 0.16706249999999997)(0.4552188552188552, 0.08701171874999998)(0.4606060606060607, 0.025523437499999996)(0.46599326599326596, 0.026683593749999995)(0.47138047138047146, 0.003480468749999999)(0.47676767676767673, 0.0011601562499999997)(0.4821548821548822, 0.0)(0.4875420875420875, 0.0)(0.492929292929293, 0.0)(0.49831649831649827, 0.0)(0.5037037037037038, 0.0011601562499999997)(0.509090909090909, 0.0)(0.5144781144781145, 0.0011601562499999997)(0.5198653198653198, 0.004640624999999999)(0.5252525252525253, 0.005800781249999999)(0.5306397306397306, 0.019722656249999995)(0.5360269360269361, 0.05568749999999999)(0.5414141414141413, 0.11485546874999998)(0.5468013468013468, 0.21694921874999995)(0.5521885521885521, 0.3921328124999999)(0.5575757575757576, 0.7250976562499999)(0.5629629629629629, 1.0998281249999997)(0.5683501683501684, 1.7425546874999998)(0.5737373737373737, 2.4896953124999994)(0.5791245791245792, 3.4166601562499994)(0.5845117845117846, 3.9688945312499992)(0.5898989898989899, 4.559414062499999)(0.5952861952861954, 4.876136718749999)(0.6006734006734007, 4.902820312499999)(0.6060606060606062, 4.515328124999999)(0.6114478114478115, 3.852878906249999)(0.616835016835017, 3.1219804687499995)(0.6222222222222222, 2.3377148437499997)(0.6276094276094277, 1.5348867187499997)(0.632996632996633, 1.0139765625)(0.6383838383838385, 0.5731171874999998)(0.6437710437710438, 0.30628124999999995)(0.6491582491582493, 0.17054296874999997)(0.6545454545454545, 0.09165234374999998)(0.65993265993266, 0.029003906249999996)(0.6653198653198653, 0.012761718749999998)(0.6707070707070708, 0.003480468749999999)(0.6760942760942761, 0.0011601562499999997)(0.6814814814814816, 0.0)(0.6868686868686869, 0.0)(0.6922558922558923, 0.0)(0.6976430976430976, 0.0)(0.7030303030303031, 0.0)(0.7084175084175084, 0.0)(0.7138047138047139, 0.0)(0.7191919191919192, 0.0)(0.7245791245791247, 0.0)(0.7299663299663299, 0.0)(0.7353535353535354, 0.0)(0.7407407407407407, 0.0)(0.7461279461279462, 0.0)(0.7515151515151515, 0.0)(0.756902356902357, 0.0)(0.7622895622895622, 0.0)(0.7676767676767677, 0.0)(0.773063973063973, 0.0)(0.7784511784511785, 0.0)(0.7838383838383838, 0.0)(0.7892255892255893, 0.0)(0.7946127946127945, 0.0)
            };
        \end{axis}
\end{tikzpicture}

%% file: tikzdiagrams/toy_data_graph_2.tex
\begin{tikzpicture}[xscale=0.46,yscale=0.46,every node/.append style={font=\Large}]
        \begin{axis}[
            ylabel near ticks, 
            xlabel={\huge$x$},
            xmin=-0.8, xmax=0.8,
            ymin=-0.2, ymax=11,
            xticklabel style={/pgf/number format/fixed},
            ticks=none,
            legend pos=north east,
            ymajorgrids=false,
        ]
        \addplot[,color=mycolour2,mark size=4pt,]
            coordinates {
               (-0.7973063973063974, 0.0)(-0.7919191919191919, 0.0)(-0.7865319865319866, 0.0)(-0.7811447811447811, 0.0)(-0.7757575757575759, 0.0)(-0.7703703703703704, 0.0)(-0.7649831649831651, 0.0)(-0.7595959595959596, 0.0)(-0.7542087542087543, 0.0)(-0.7488215488215488, 0.0)(-0.7434343434343436, 0.0)(-0.7380471380471381, 0.0)(-0.7326599326599328, 0.0)(-0.7272727272727273, 0.0)(-0.721885521885522, 0.0)(-0.7164983164983165, 0.0)(-0.7111111111111112, 0.0)(-0.7057239057239058, 0.0)(-0.7003367003367004, 0.0)(-0.694949494949495, 0.0)(-0.6895622895622896, 0.0)(-0.6841750841750842, 0.0)(-0.6787878787878788, 0.0)(-0.6734006734006734, 0.0)(-0.6680134680134681, 0.0)(-0.6626262626262627, 0.0)(-0.6572390572390573, 0.0)(-0.6518518518518519, 0.0)(-0.6464646464646464, 0.0)(-0.6410774410774411, 0.0)(-0.6356902356902356, 0.0)(-0.6303030303030304, 0.0)(-0.6249158249158249, 0.0)(-0.6195286195286196, 0.0)(-0.6141414141414141, 0.0)(-0.6087542087542088, 0.0)(-0.6033670033670033, 0.0)(-0.5979797979797981, 0.0)(-0.5925925925925926, 0.0)(-0.5872053872053873, 0.0)(-0.5818181818181818, 0.0)(-0.5764309764309765, 0.0)(-0.571043771043771, 0.0)(-0.5656565656565657, 0.0)(-0.5602693602693603, 0.0)(-0.554882154882155, 0.0)(-0.5494949494949495, 0.0)(-0.5441077441077442, 0.0)(-0.5387205387205387, 0.0)(-0.5333333333333334, 0.0)(-0.5279461279461279, 0.0)(-0.5225589225589227, 0.0)(-0.5171717171717172, 0.0)(-0.5117845117845118, 0.0)(-0.5063973063973064, 0.0)(-0.501010101010101, 0.0)(-0.49562289562289563, 0.0)(-0.49023569023569025, 0.0)(-0.48484848484848486, 0.0)(-0.4794612794612795, 0.0)(-0.4740740740740741, 0.0)(-0.4686868686868687, 0.0)(-0.4632996632996633, 0.028031249999999997)(-0.45791245791245794, 0.06073437499999999)(-0.45252525252525255, 0.058865625)(-0.44713804713804717, 0.19621875)(-0.4417508417508418, 0.47185937499999997)(-0.4363636363636364, 1.05210625)(-0.430976430976431, 1.5445218749999998)(-0.42558922558922563, 2.84236875)(-0.42020202020202024, 5.673525)(-0.41481481481481486, 7.01341875)(-0.4094276094276095, 9.0858625)(-0.4040404040404041, 9.398878125)(-0.3986531986531987, 10.34353125)(-0.39326599326599326, 8.836384375)(-0.3878787878787879, 7.409593749999999)(-0.3824915824915825, 4.641040625)(-0.3771043771043771, 2.58821875)(-0.3717171717171717, 1.7818531249999998)(-0.36632996632996634, 1.3641874999999999)(-0.36094276094276095, 0.3401125)(-0.35555555555555557, 0.1457625)(-0.3501683501683502, 0.096240625)(-0.3447811447811448, 0.0)(-0.3393939393939394, 0.0)(-0.33400673400673403, 0.0)(-0.32861952861952864, 0.0)(-0.32323232323232326, 0.0)(-0.3178451178451179, 0.0)(-0.3124579124579125, 0.0)(-0.3070707070707071, 0.0)(-0.3016835016835017, 0.0)(-0.29629629629629634, 0.0)(-0.29090909090909095, 0.0)(-0.28552188552188557, 0.0)(-0.2801346801346802, 0.0)(-0.2747474747474748, 0.0)(-0.2693602693602694, 0.0)(-0.263973063973064, 0.0)(-0.25858585858585864, 0.0)(-0.25319865319865326, 0.0)(-0.24781144781144787, 0.0)(-0.2424242424242425, 0.0)(-0.2370370370370371, 0.0)(-0.23164983164983172, 0.0)(-0.22626262626262628, 0.0)(-0.22087542087542084, 0.0)(-0.21548821548821545, 0.0)(-0.21010101010101007, 0.0)(-0.20471380471380468, 0.0)(-0.1993265993265993, 0.0)(-0.1939393939393939, 0.0)(-0.18855218855218853, 0.0)(-0.18316498316498314, 0.0)(-0.17777777777777776, 0.0)(-0.17239057239057237, 0.0)(-0.167003367003367, 0.0)(-0.1616161616161616, 0.0)(-0.15622895622895622, 0.0)(-0.15084175084175083, 0.0)(-0.14545454545454545, 0.0)(-0.14006734006734006, 0.0)(-0.13468013468013468, 0.0)(-0.1292929292929293, 0.0)(-0.12390572390572391, 0.0)(-0.11851851851851852, 0.0)(-0.11313131313131314, 0.0)(-0.10774410774410775, 0.0)(-0.10235690235690237, 0.0)(-0.09696969696969698, 0.0)(-0.0915824915824916, 0.0)(-0.08619528619528621, 0.0)(-0.08080808080808083, 0.0)(-0.07542087542087544, 0.0)(-0.07003367003367006, 0.0)(-0.06464646464646467, 0.0)(-0.05925925925925929, 0.0)(-0.053872053872053904, 0.0)(-0.04848484848484852, 0.0)(-0.043097643097643135, 0.0)(-0.03771043771043775, 0.0)(-0.032323232323232365, 0.0)(-0.02693602693602698, 0.0)(-0.021548821548821595, 0.0)(-0.01616161616161621, 0.0)(-0.010774410774410825, 0.0)(-0.00538720538720544, 0.0)(-5.551115123125783e-17, 0.0)(0.005387205387205385, 0.0)(0.010774410774410825, 0.0)(0.01616161616161621, 0.0)(0.021548821548821595, 0.0)(0.02693602693602698, 0.0)(0.032323232323232365, 0.0)(0.03771043771043775, 0.0)(0.043097643097643135, 0.0)(0.04848484848484852, 0.0)(0.053872053872053904, 0.0)(0.05925925925925929, 0.0)(0.06464646464646467, 0.0)(0.07003367003367006, 0.0)(0.07542087542087544, 0.0)(0.08080808080808083, 0.0)(0.08619528619528621, 0.0)(0.0915824915824916, 0.0)(0.09696969696969698, 0.0)(0.10235690235690237, 0.0)(0.10774410774410775, 0.0)(0.11313131313131314, 0.0)(0.11851851851851852, 0.0)(0.12390572390572391, 0.0)(0.1292929292929293, 0.0)(0.13468013468013468, 0.0)(0.14006734006734006, 0.0)(0.14545454545454545, 0.0)(0.15084175084175083, 0.0)(0.15622895622895622, 0.0)(0.1616161616161616, 0.0)(0.167003367003367, 0.0)(0.17239057239057237, 0.0)(0.17777777777777776, 0.0)(0.18316498316498314, 0.0)(0.18855218855218853, 0.0)(0.1939393939393939, 0.0)(0.1993265993265993, 0.0)(0.20471380471380474, 0.0)(0.21010101010101012, 0.0)(0.2154882154882155, 0.0)(0.2208754208754209, 0.0)(0.22626262626262628, 0.0)(0.23164983164983166, 0.0)(0.23703703703703705, 0.0)(0.24242424242424243, 0.0)(0.24781144781144782, 0.0)(0.2531986531986532, 0.0)(0.2585858585858586, 0.0)(0.26397306397306397, 0.0)(0.26936026936026936, 0.0)(0.27474747474747474, 0.0)(0.2801346801346801, 0.0)(0.2855218855218855, 0.0)(0.2909090909090909, 0.0)(0.2962962962962963, 0.0)(0.30168350168350166, 0.0)(0.30707070707070705, 0.0)(0.31245791245791243, 0.0)(0.3178451178451178, 0.0)(0.3232323232323232, 0.0)(0.3286195286195286, 0.0)(0.334006734006734, 0.0)(0.33939393939393936, 0.0)(0.34478114478114474, 0.03176875)(0.35016835016835024, 0.058865625)(0.3555555555555556, 0.08783125)(0.360942760942761, 0.262559375)(0.3663299663299664, 0.5680999999999999)(0.3717171717171718, 1.3118625)(0.37710437710437716, 1.400628125)(0.38249158249158255, 3.536609375)(0.38787878787878793, 5.853859375)(0.3932659932659933, 7.749706249999999)(0.3986531986531987, 9.129778125)(0.4040404040404041, 9.7436625)(0.4094276094276095, 9.497921875)(0.41481481481481486, 9.183971875)(0.42020202020202024, 6.6377999999999995)(0.42558922558922563, 4.091628125)(0.430976430976431, 2.179896875)(0.4363636363636364, 1.7753124999999998)(0.4417508417508418, 0.961471875)(0.44713804713804717, 0.25601874999999996)(0.45252525252525255, 0.20743124999999998)(0.45791245791245794, 0.0)(0.4632996632996633, 0.0)(0.4686868686868687, 0.0)(0.4740740740740741, 0.0)(0.4794612794612795, 0.0)(0.48484848484848486, 0.0)(0.49023569023569025, 0.0)(0.49562289562289563, 0.0)(0.501010101010101, 0.0)(0.5063973063973064, 0.0)(0.5117845117845118, 0.0)(0.5171717171717172, 0.0)(0.5225589225589226, 0.0)(0.5279461279461279, 0.0)(0.5333333333333333, 0.0)(0.5387205387205387, 0.0)(0.5441077441077441, 0.0)(0.5494949494949495, 0.0)(0.5548821548821549, 0.0)(0.5602693602693603, 0.0)(0.5656565656565656, 0.0)(0.571043771043771, 0.0)(0.5764309764309764, 0.0)(0.5818181818181819, 0.0)(0.5872053872053873, 0.0)(0.5925925925925927, 0.0)(0.5979797979797981, 0.0)(0.6033670033670034, 0.0)(0.6087542087542088, 0.0)(0.6141414141414142, 0.0)(0.6195286195286196, 0.0)(0.624915824915825, 0.0)(0.6303030303030304, 0.0)(0.6356902356902357, 0.0)(0.6410774410774411, 0.0)(0.6464646464646465, 0.0)(0.6518518518518519, 0.0)(0.6572390572390573, 0.0)(0.6626262626262627, 0.0)(0.6680134680134681, 0.0)(0.6734006734006734, 0.0)(0.6787878787878788, 0.0)(0.6841750841750842, 0.0)(0.6895622895622896, 0.0)(0.694949494949495, 0.0)(0.7003367003367004, 0.0)(0.7057239057239058, 0.0)(0.7111111111111111, 0.0)(0.7164983164983165, 0.0)(0.7218855218855219, 0.0)(0.7272727272727273, 0.0)(0.7326599326599327, 0.0)(0.7380471380471381, 0.0)(0.7434343434343434, 0.0)(0.7488215488215488, 0.0)(0.7542087542087542, 0.0)(0.7595959595959596, 0.0)(0.764983164983165, 0.0)(0.7703703703703704, 0.0)(0.7757575757575758, 0.0)(0.7811447811447811, 0.0)(0.7865319865319865, 0.0)(0.7919191919191919, 0.0)(0.7973063973063973, 0.0)
            };
        \addplot[,color=mycolour3,mark size=4pt,]
            coordinates {
               (-0.7973063973063974, 0.0)(-0.7919191919191919, 0.0)(-0.7865319865319866, 0.0)(-0.7811447811447811, 0.0)(-0.7757575757575759, 0.0)(-0.7703703703703704, 0.0)(-0.7649831649831651, 0.0)(-0.7595959595959596, 0.0)(-0.7542087542087543, 0.0)(-0.7488215488215488, 0.0)(-0.7434343434343436, 0.0)(-0.7380471380471381, 0.0)(-0.7326599326599328, 0.0)(-0.7272727272727273, 0.0)(-0.721885521885522, 0.0)(-0.7164983164983165, 0.0)(-0.7111111111111112, 0.0)(-0.7057239057239058, 0.0)(-0.7003367003367004, 0.0)(-0.694949494949495, 0.0)(-0.6895622895622896, 0.0)(-0.6841750841750842, 0.0)(-0.6787878787878788, 0.0)(-0.6734006734006734, 0.0)(-0.6680134680134681, 0.0)(-0.6626262626262627, 0.0)(-0.6572390572390573, 0.0)(-0.6518518518518519, 0.0)(-0.6464646464646464, 0.0)(-0.6410774410774411, 0.0)(-0.6356902356902356, 0.0)(-0.6303030303030304, 0.0)(-0.6249158249158249, 0.0)(-0.6195286195286196, 0.0)(-0.6141414141414141, 0.0)(-0.6087542087542088, 0.0)(-0.6033670033670033, 0.0)(-0.5979797979797981, 0.0)(-0.5925925925925926, 0.0)(-0.5872053872053873, 0.0)(-0.5818181818181818, 0.0)(-0.5764309764309765, 0.0)(-0.571043771043771, 0.0)(-0.5656565656565657, 0.0)(-0.5602693602693603, 0.0)(-0.554882154882155, 0.0)(-0.5494949494949495, 0.0)(-0.5441077441077442, 0.0)(-0.5387205387205387, 0.0)(-0.5333333333333334, 0.04952187500000001)(-0.5279461279461279, 0.08409375000000001)(-0.5225589225589227, 0.12053437500000001)(-0.5171717171717172, 0.23546250000000002)(-0.5117845117845118, 0.5353968750000001)(-0.5063973063973064, 0.9596031250000001)(-0.501010101010101, 1.8173593750000003)(-0.49562289562289563, 2.5452375000000003)(-0.49023569023569025, 3.8916718750000006)(-0.48484848484848486, 4.691496875)(-0.4794612794612795, 4.9811531250000005)(-0.4740740740740741, 4.796146875000001)(-0.4686868686868687, 4.236456250000001)(-0.4632996632996633, 3.1310906250000006)(-0.45791245791245794, 2.1499968750000003)(-0.45252525252525255, 1.5594718750000003)(-0.44713804713804717, 0.8381343750000001)(-0.4417508417508418, 0.5783781250000001)(-0.4363636363636364, 0.13922187500000002)(-0.430976430976431, 0.20836562500000003)(-0.42558922558922563, 0.031768750000000005)(-0.42020202020202024, 0.0009343750000000001)(-0.41481481481481486, 0.0)(-0.4094276094276095, 0.0)(-0.4040404040404041, 0.0)(-0.3986531986531987, 0.0)(-0.39326599326599326, 0.0018687500000000002)(-0.3878787878787879, 0.0)(-0.3824915824915825, 0.0)(-0.3771043771043771, 0.0018687500000000002)(-0.3717171717171717, 0.028965625000000005)(-0.36632996632996634, 0.042046875000000004)(-0.36094276094276095, 0.10932187500000001)(-0.35555555555555557, 0.292459375)(-0.3501683501683502, 0.36347187500000006)(-0.3447811447811448, 0.7278781250000002)(-0.3393939393939394, 1.4660343750000002)(-0.33400673400673403, 2.29669375)(-0.32861952861952864, 3.3441281250000006)(-0.32323232323232326, 4.508359375)(-0.3178451178451179, 5.004512500000001)(-0.3124579124579125, 4.540128125000001)(-0.3070707070707071, 4.553209375000001)(-0.3016835016835017, 3.9038187500000006)(-0.29629629629629634, 2.3957375000000005)(-0.29090909090909095, 1.5781593750000003)(-0.28552188552188557, 1.026878125)(-0.2801346801346802, 0.6998468750000001)(-0.2747474747474748, 0.21584062500000004)(-0.2693602693602694, 0.23359375000000004)(-0.263973063973064, 0.05699687500000001)(-0.25858585858585864, 0.0)(-0.25319865319865326, 0.0)(-0.24781144781144787, 0.0)(-0.2424242424242425, 0.0)(-0.2370370370370371, 0.0)(-0.23164983164983172, 0.0)(-0.22626262626262628, 0.0)(-0.22087542087542084, 0.0)(-0.21548821548821545, 0.0)(-0.21010101010101007, 0.0)(-0.20471380471380468, 0.0)(-0.1993265993265993, 0.0)(-0.1939393939393939, 0.0)(-0.18855218855218853, 0.0)(-0.18316498316498314, 0.0)(-0.17777777777777776, 0.0)(-0.17239057239057237, 0.0)(-0.167003367003367, 0.0)(-0.1616161616161616, 0.0)(-0.15622895622895622, 0.0)(-0.15084175084175083, 0.0)(-0.14545454545454545, 0.0)(-0.14006734006734006, 0.0)(-0.13468013468013468, 0.0)(-0.1292929292929293, 0.0)(-0.12390572390572391, 0.0)(-0.11851851851851852, 0.0)(-0.11313131313131314, 0.0)(-0.10774410774410775, 0.0)(-0.10235690235690237, 0.0)(-0.09696969696969698, 0.0)(-0.0915824915824916, 0.0)(-0.08619528619528621, 0.0)(-0.08080808080808083, 0.0)(-0.07542087542087544, 0.0)(-0.07003367003367006, 0.0)(-0.06464646464646467, 0.0)(-0.05925925925925929, 0.0)(-0.053872053872053904, 0.0)(-0.04848484848484852, 0.0)(-0.043097643097643135, 0.0)(-0.03771043771043775, 0.0)(-0.032323232323232365, 0.0)(-0.02693602693602698, 0.0)(-0.021548821548821595, 0.0)(-0.01616161616161621, 0.0)(-0.010774410774410825, 0.0)(-0.00538720538720544, 0.0)(-5.551115123125783e-17, 0.0)(0.005387205387205385, 0.0)(0.010774410774410825, 0.0)(0.01616161616161621, 0.0)(0.021548821548821595, 0.0)(0.02693602693602698, 0.0)(0.032323232323232365, 0.0)(0.03771043771043775, 0.0)(0.043097643097643135, 0.0)(0.04848484848484852, 0.0)(0.053872053872053904, 0.0)(0.05925925925925929, 0.0)(0.06464646464646467, 0.0)(0.07003367003367006, 0.0)(0.07542087542087544, 0.0)(0.08080808080808083, 0.0)(0.08619528619528621, 0.0)(0.0915824915824916, 0.0)(0.09696969696969698, 0.0)(0.10235690235690237, 0.0)(0.10774410774410775, 0.0)(0.11313131313131314, 0.0)(0.11851851851851852, 0.0)(0.12390572390572391, 0.0)(0.1292929292929293, 0.0)(0.13468013468013468, 0.0)(0.14006734006734006, 0.0)(0.14545454545454545, 0.0)(0.15084175084175083, 0.0)(0.15622895622895622, 0.0)(0.1616161616161616, 0.0)(0.167003367003367, 0.0)(0.17239057239057237, 0.0)(0.17777777777777776, 0.0)(0.18316498316498314, 0.0)(0.18855218855218853, 0.0)(0.1939393939393939, 0.0)(0.1993265993265993, 0.0)(0.20471380471380474, 0.0)(0.21010101010101012, 0.0)(0.2154882154882155, 0.0)(0.2208754208754209, 0.0)(0.22626262626262628, 0.0)(0.23164983164983166, 0.0)(0.23703703703703705, 0.0)(0.24242424242424243, 0.0)(0.24781144781144782, 0.0)(0.2531986531986532, 0.0)(0.2585858585858586, 0.0)(0.26397306397306397, 0.0)(0.26936026936026936, 0.0)(0.27474747474747474, 0.014950000000000001)(0.2801346801346801, 0.06447187500000001)(0.2855218855218855, 0.10091250000000002)(0.2909090909090909, 0.21770937500000004)(0.2962962962962963, 0.40084687500000005)(0.30168350168350166, 0.6409812500000001)(0.30707070707070705, 1.2782250000000002)(0.31245791245791243, 2.3480843750000004)(0.3178451178451178, 3.0955843750000005)(0.3232323232323232, 4.504621875000001)(0.3286195286195286, 5.048428125000001)(0.334006734006734, 4.322418750000001)(0.33939393939393936, 4.590584375000001)(0.34478114478114474, 3.9813718750000007)(0.35016835016835024, 2.5695312500000003)(0.3555555555555556, 1.5445218750000003)(0.360942760942761, 1.273553125)(0.3663299663299664, 0.662471875)(0.3717171717171718, 0.23826562500000004)(0.37710437710437716, 0.14669687500000003)(0.38249158249158255, 0.10651875000000001)(0.38787878787878793, 0.014950000000000001)(0.3932659932659933, 0.0)(0.3986531986531987, 0.0009343750000000001)(0.4040404040404041, 0.0)(0.4094276094276095, 0.0009343750000000001)(0.41481481481481486, 0.0)(0.42020202020202024, 0.0009343750000000001)(0.42558922558922563, 0.0)(0.430976430976431, 0.017753125)(0.4363636363636364, 0.047653125000000005)(0.4417508417508418, 0.08502812500000001)(0.44713804713804717, 0.20182500000000003)(0.45252525252525255, 0.36720937500000006)(0.45791245791245794, 0.6578)(0.4632996632996633, 1.2875687500000002)(0.4686868686868687, 2.31538125)(0.4740740740740741, 3.1946281250000004)(0.4794612794612795, 4.1177906250000005)(0.48484848484848486, 5.026937500000001)(0.49023569023569025, 4.5064906250000005)(0.49562289562289563, 4.633565625000001)(0.501010101010101, 4.237390625000001)(0.5063973063973064, 2.4461937500000004)(0.5117845117845118, 1.6230093750000003)(0.5171717171717172, 1.2754218750000001)(0.5225589225589226, 0.7334843750000001)(0.5279461279461279, 0.27377187500000005)(0.5333333333333333, 0.18594062500000003)(0.5387205387205387, 0.10371562500000002)(0.5441077441077441, 0.019621875000000004)(0.5494949494949495, 0.0)(0.5548821548821549, 0.0)(0.5602693602693603, 0.0)(0.5656565656565656, 0.0)(0.571043771043771, 0.0)(0.5764309764309764, 0.0)(0.5818181818181819, 0.0)(0.5872053872053873, 0.0)(0.5925925925925927, 0.0)(0.5979797979797981, 0.0)(0.6033670033670034, 0.0)(0.6087542087542088, 0.0)(0.6141414141414142, 0.0)(0.6195286195286196, 0.0)(0.624915824915825, 0.0)(0.6303030303030304, 0.0)(0.6356902356902357, 0.0)(0.6410774410774411, 0.0)(0.6464646464646465, 0.0)(0.6518518518518519, 0.0)(0.6572390572390573, 0.0)(0.6626262626262627, 0.0)(0.6680134680134681, 0.0)(0.6734006734006734, 0.0)(0.6787878787878788, 0.0)(0.6841750841750842, 0.0)(0.6895622895622896, 0.0)(0.694949494949495, 0.0)(0.7003367003367004, 0.0)(0.7057239057239058, 0.0)(0.7111111111111111, 0.0)(0.7164983164983165, 0.0)(0.7218855218855219, 0.0)(0.7272727272727273, 0.0)(0.7326599326599327, 0.0)(0.7380471380471381, 0.0)(0.7434343434343434, 0.0)(0.7488215488215488, 0.0)(0.7542087542087542, 0.0)(0.7595959595959596, 0.0)(0.764983164983165, 0.0)(0.7703703703703704, 0.0)(0.7757575757575758, 0.0)(0.7811447811447811, 0.0)(0.7865319865319865, 0.0)(0.7919191919191919, 0.0)(0.7973063973063973, 0.0)
            };
        \legend{2-code VQ-VAE, 4-code VQ-VAE}
        \end{axis}
\end{tikzpicture}

%% file: tikzdiagrams/toy_data_graph_4.tex
\begin{tikzpicture}[xscale=0.46,yscale=0.46,every node/.append style={font=\Large}]
        \begin{axis}[
            ylabel near ticks,
            xlabel={\huge$x$},
            xmin=-0.8, xmax=0.8,
            ymin=-0.2, ymax=11,
            xticklabel style={/pgf/number format/fixed},
            ticks=none,
            legend pos=north east,
            ymajorgrids=false,
        ]
        \addplot[,color=mycolour3,mark size=4pt,]
            coordinates {
               (-0.9966555183946488, 0.0)(-0.9899665551839465, 0.0)(-0.9832775919732442, 0.0)(-0.9765886287625418, 0.0)(-0.9698996655518395, 0.0)(-0.9632107023411371, 0.0)(-0.9565217391304348, 0.0)(-0.9498327759197325, 0.0)(-0.9431438127090301, 0.0)(-0.9364548494983278, 0.0)(-0.9297658862876255, 0.0)(-0.9230769230769231, 0.0)(-0.9163879598662208, 0.0)(-0.9096989966555185, 0.0)(-0.9030100334448161, 0.0)(-0.8963210702341137, 0.0)(-0.8896321070234113, 0.0)(-0.882943143812709, 0.0)(-0.8762541806020067, 0.0)(-0.8695652173913043, 0.0)(-0.862876254180602, 0.0)(-0.8561872909698997, 0.0)(-0.8494983277591973, 0.0)(-0.842809364548495, 0.0)(-0.8361204013377926, 0.0)(-0.8294314381270903, 0.0)(-0.822742474916388, 0.0)(-0.8160535117056856, 0.0)(-0.8093645484949833, 0.0)(-0.802675585284281, 0.0)(-0.7959866220735786, 0.0)(-0.7892976588628763, 0.0)(-0.782608695652174, 0.0)(-0.7759197324414716, 0.0)(-0.7692307692307693, 0.0)(-0.7625418060200668, 0.0)(-0.7558528428093645, 0.0)(-0.7491638795986622, 0.0)(-0.7424749163879598, 0.0)(-0.7357859531772575, 0.0)(-0.7290969899665551, 0.0)(-0.7224080267558528, 0.0)(-0.7157190635451505, 0.0)(-0.7090301003344481, 0.0)(-0.7023411371237458, 0.0)(-0.6956521739130435, 0.0)(-0.6889632107023411, 0.0)(-0.6822742474916388, 0.0)(-0.6755852842809364, 0.0)(-0.6688963210702341, 0.0)(-0.6622073578595318, 0.06260312500000004)(-0.6555183946488294, 0.4045843750000002)(-0.6488294314381271, 0.5681000000000003)(-0.6421404682274248, 1.5370468750000008)(-0.6354515050167224, 2.046281250000001)(-0.6287625418060201, 4.381284375000003)(-0.6220735785953178, 5.507206250000003)(-0.6153846153846154, 7.801096875000004)(-0.6086956521739131, 8.330887500000005)(-0.6020066889632107, 10.222996875000005)(-0.5953177257525084, 9.289556250000006)(-0.5886287625418061, 8.282300000000005)(-0.5819397993311037, 6.1575312500000035)(-0.5752508361204014, 4.9662031250000025)(-0.5685618729096991, 2.429375000000001)(-0.5618729096989967, 1.3184031250000008)(-0.5551839464882944, 0.7531062500000004)(-0.548494983277592, 0.41392812500000026)(-0.5418060200668897, 0.1775312500000001)(-0.5351170568561874, 0.030834375000000018)(-0.528428093645485, 0.025228125000000014)(-0.5217391304347826, 0.0)(-0.5150501672240803, 0.0)(-0.5083612040133779, 0.0)(-0.5016722408026756, 0.0)(-0.49498327759197325, 0.0)(-0.4882943143812709, 0.0)(-0.4816053511705686, 0.0)(-0.47491638795986624, 0.0)(-0.4682274247491639, 0.0)(-0.46153846153846156, 0.0)(-0.4548494983277592, 0.0)(-0.4481605351170569, 0.0)(-0.44147157190635455, 0.0)(-0.4347826086956522, 0.0)(-0.4280936454849499, 0.0)(-0.42140468227424754, 0.0)(-0.4147157190635452, 0.0)(-0.40802675585284287, 0.0)(-0.4013377926421405, 0.0)(-0.3946488294314381, 0.0)(-0.38795986622073575, 0.0)(-0.3812709030100334, 0.0)(-0.3745819397993311, 0.0)(-0.36789297658862874, 0.0)(-0.3612040133779264, 0.0)(-0.35451505016722407, 0.0)(-0.34782608695652173, 0.0)(-0.3411371237458194, 0.0)(-0.33444816053511706, 0.0)(-0.3277591973244147, 0.0)(-0.3210702341137124, 0.0)(-0.31438127090301005, 0.0)(-0.3076923076923077, 0.0)(-0.3010033444816054, 0.0)(-0.29431438127090304, 0.0)(-0.2876254180602007, 0.0)(-0.28093645484949836, 0.0)(-0.274247491638796, 0.0)(-0.2675585284280937, 0.0)(-0.26086956521739135, 0.0)(-0.254180602006689, 0.0)(-0.24749163879598668, 0.0)(-0.24080267558528434, 0.0)(-0.234113712374582, 0.0)(-0.2274247491638796, 0.0)(-0.22073578595317722, 0.0)(-0.21404682274247488, 0.0)(-0.20735785953177255, 0.0)(-0.2006688963210702, 0.0)(-0.19397993311036787, 0.0)(-0.18729096989966554, 0.0)(-0.1806020066889632, 0.0)(-0.17391304347826086, 0.0)(-0.16722408026755853, 0.0)(-0.1605351170568562, 0.0)(-0.15384615384615385, 0.0)(-0.14715719063545152, 0.0)(-0.14046822742474918, 0.0)(-0.13377926421404684, 0.0)(-0.1270903010033445, 0.0)(-0.12040133779264217, 0.0)(-0.11371237458193983, 0.0)(-0.1070234113712375, 0.0)(-0.10033444816053516, 0.0)(-0.09364548494983282, 0.0)(-0.08695652173913049, 0.0)(-0.08026755852842815, 0.0)(-0.07357859531772581, 0.0)(-0.06688963210702348, 0.0)(-0.060200668896321086, 0.0)(-0.05351170568561869, 0.0)(-0.04682274247491636, 0.0)(-0.04013377926421402, 0.0)(-0.03344481605351168, 0.0)(-0.026755852842809347, 0.0)(-0.02006688963210701, 0.0)(-0.013377926421404673, 0.0)(-0.006688963210702337, 0.0)(0.0, 0.0)(0.006688963210702337, 0.0)(0.013377926421404673, 0.0)(0.02006688963210701, 0.0)(0.026755852842809347, 0.0)(0.03344481605351168, 0.0)(0.04013377926421402, 0.0)(0.04682274247491636, 0.0)(0.05351170568561869, 0.0)(0.06020066889632103, 0.0)(0.06688963210702337, 0.0)(0.0735785953177257, 0.0)(0.08026755852842804, 0.0)(0.08695652173913038, 0.0)(0.09364548494983271, 0.0)(0.10033444816053505, 0.0)(0.10702341137123739, 0.0)(0.11371237458193972, 0.0)(0.12040133779264206, 0.0)(0.1270903010033444, 0.0)(0.13377926421404673, 0.0)(0.14046822742474907, 0.0)(0.1471571906354514, 0.0)(0.15384615384615374, 0.0)(0.16053511705685608, 0.0)(0.16722408026755842, 0.0)(0.17391304347826075, 0.0)(0.1806020066889631, 0.0)(0.18729096989966543, 0.0)(0.19397993311036776, 0.0)(0.2006688963210702, 0.0)(0.20735785953177266, 0.0)(0.214046822742475, 0.0)(0.22073578595317733, 0.0)(0.22742474916387967, 0.0)(0.234113712374582, 0.0)(0.24080267558528434, 0.0)(0.24749163879598668, 0.0)(0.254180602006689, 0.0)(0.26086956521739135, 0.0)(0.2675585284280937, 0.0)(0.274247491638796, 0.0)(0.28093645484949836, 0.0)(0.2876254180602007, 0.0)(0.29431438127090304, 0.0)(0.3010033444816054, 0.0)(0.3076923076923077, 0.0)(0.31438127090301005, 0.0)(0.3210702341137124, 0.0)(0.3277591973244147, 0.0)(0.33444816053511706, 0.0)(0.3411371237458194, 0.0)(0.34782608695652173, 0.0)(0.35451505016722407, 0.0)(0.3612040133779264, 0.0)(0.36789297658862874, 0.0)(0.3745819397993311, 0.0)(0.3812709030100334, 0.0)(0.38795986622073575, 0.0)(0.3946488294314381, 0.0)(0.4013377926421404, 0.0)(0.40802675585284276, 0.0)(0.4147157190635451, 0.0)(0.42140468227424743, 0.0)(0.42809364548494977, 0.0)(0.4347826086956521, 0.0)(0.44147157190635444, 0.0)(0.4481605351170568, 0.0)(0.4548494983277591, 0.0)(0.46153846153846145, 0.0)(0.4682274247491638, 0.0)(0.4749163879598661, 0.0)(0.48160535117056846, 0.0)(0.4882943143812708, 0.0)(0.49498327759197314, 0.0)(0.5016722408026755, 0.0)(0.5083612040133778, 0.0)(0.5150501672240801, 0.0)(0.5217391304347825, 0.0)(0.5284280936454848, 0.0)(0.5351170568561872, 0.03176875000000002)(0.5418060200668896, 0.19154687500000012)(0.548494983277592, 0.31862187500000017)(0.5551839464882944, 1.0904156250000006)(0.5618729096989967, 1.6398281250000009)(0.5685618729096991, 3.589868750000002)(0.5752508361204014, 4.845668750000002)(0.5819397993311037, 6.3032937500000035)(0.5886287625418061, 8.349575000000005)(0.5953177257525084, 9.511937500000005)(0.6020066889632107, 9.656765625000006)(0.6086956521739131, 9.420368750000005)(0.6153846153846154, 6.638734375000004)(0.6220735785953178, 5.976262500000003)(0.6287625418060201, 3.331981250000002)(0.6354515050167224, 1.833243750000001)(0.6421404682274248, 1.1006937500000007)(0.6488294314381271, 0.5456750000000004)(0.6555183946488294, 0.29432812500000016)(0.6622073578595318, 0.059800000000000034)(0.6688963210702341, 0.028965625000000016)(0.6755852842809364, 0.03457187500000002)(0.6822742474916388, 0.0)(0.6889632107023411, 0.0)(0.6956521739130435, 0.0)(0.7023411371237458, 0.0)(0.7090301003344481, 0.0)(0.7157190635451505, 0.0)(0.7224080267558528, 0.0)(0.7290969899665551, 0.0)(0.7357859531772575, 0.0)(0.7424749163879598, 0.0)(0.7491638795986622, 0.0)(0.7558528428093645, 0.0)(0.7625418060200668, 0.0)(0.7692307692307692, 0.0)(0.7759197324414715, 0.0)(0.7826086956521738, 0.0)(0.7892976588628762, 0.0)(0.7959866220735785, 0.0)(0.8026755852842808, 0.0)(0.8093645484949832, 0.0)(0.8160535117056855, 0.0)(0.8227424749163879, 0.0)(0.8294314381270902, 0.0)(0.8361204013377925, 0.0)(0.8428093645484949, 0.0)(0.8494983277591972, 0.0)(0.8561872909698995, 0.0)(0.8628762541806019, 0.0)(0.8695652173913042, 0.0)(0.8762541806020065, 0.0)(0.882943143812709, 0.0)(0.8896321070234114, 0.0)(0.8963210702341138, 0.0)(0.9030100334448161, 0.0)(0.9096989966555185, 0.0)(0.9163879598662208, 0.0)(0.9230769230769231, 0.0)(0.9297658862876255, 0.0)(0.9364548494983278, 0.0)(0.9431438127090301, 0.0)(0.9498327759197325, 0.0)(0.9565217391304348, 0.0)(0.9632107023411371, 0.0)(0.9698996655518395, 0.0)(0.9765886287625418, 0.0)(0.9832775919732442, 0.0)(0.9899665551839465, 0.0)(0.9966555183946488, 0.0)
            };
            \addplot[,color=mycolour1,mark size=4pt,]
            coordinates {
               (-0.8, 0.0)(-0.7946127946127947, 0.0)(-0.7892255892255893, 0.0)(-0.7838383838383839, 0.0)(-0.7784511784511785, 0.0)(-0.7730639730639731, 0.0)(-0.7676767676767677, 0.0)(-0.7622895622895624, 0.0)(-0.756902356902357, 0.0)(-0.7515151515151516, 0.0)(-0.7461279461279462, 0.0)(-0.7407407407407408, 0.0)(-0.7353535353535354, 0.0)(-0.72996632996633, 0.0)(-0.7245791245791247, 0.0)(-0.7191919191919193, 0.0)(-0.7138047138047139, 0.0)(-0.7084175084175085, 0.0)(-0.7030303030303031, 0.0011601562499999997)(-0.6976430976430976, 0.0)(-0.6922558922558923, 0.0)(-0.6868686868686869, 0.0)(-0.6814814814814816, 0.003480468749999999)(-0.6760942760942761, 0.010441406249999998)(-0.6707070707070708, 0.015082031249999997)(-0.6653198653198653, 0.041765624999999994)(-0.65993265993266, 0.08701171874999998)(-0.6545454545454545, 0.15894140624999997)(-0.6491582491582492, 0.33412499999999995)(-0.6437710437710438, 0.6380859374999999)(-0.6383838383838384, 1.0186171874999999)(-0.632996632996633, 1.4815195312499998)(-0.6276094276094276, 2.3075507812499993)(-0.6222222222222222, 3.1208203124999994)(-0.6168350168350168, 3.859839843749999)(-0.6114478114478115, 4.475882812499999)(-0.6060606060606061, 4.868015624999999)(-0.6006734006734007, 5.022316406249999)(-0.5952861952861953, 4.630183593749999)(-0.5898989898989899, 4.038503906249999)(-0.5845117845117845, 3.2832421874999995)(-0.5791245791245792, 2.5175390624999996)(-0.5737373737373738, 1.7205117187499996)(-0.5683501683501684, 1.1996015624999998)(-0.562962962962963, 0.6868124999999998)(-0.5575757575757576, 0.41997656249999993)(-0.5521885521885522, 0.20998828124999996)(-0.5468013468013468, 0.11253515624999998)(-0.5414141414141415, 0.04640624999999999)(-0.5360269360269361, 0.017402343749999997)(-0.5306397306397307, 0.011601562499999997)(-0.5252525252525253, 0.0011601562499999997)(-0.5198653198653199, 0.0)(-0.5144781144781145, 0.0)(-0.509090909090909, 0.0)(-0.5037037037037038, 0.0)(-0.4983164983164983, 0.0)(-0.49292929292929294, 0.0011601562499999997)(-0.48754208754208755, 0.0011601562499999997)(-0.48215488215488217, 0.0023203124999999995)(-0.4767676767676768, 0.0011601562499999997)(-0.4713804713804714, 0.011601562499999997)(-0.465993265993266, 0.026683593749999995)(-0.46060606060606063, 0.07076953124999999)(-0.45521885521885525, 0.16242187499999997)(-0.44983164983164986, 0.27611718749999997)(-0.4444444444444445, 0.5394726562499998)(-0.4390572390572391, 1.0186171874999999)(-0.4336700336700337, 1.5070429687499998)(-0.4282828282828283, 2.2274999999999996)(-0.42289562289562294, 3.0778945312499992)(-0.41750841750841755, 3.6753749999999994)(-0.41212121212121217, 4.381910156249999)(-0.4067340067340068, 4.764761718749999)(-0.4013468013468014, 5.060601562499999)(-0.39595959595959596, 4.692832031249999)(-0.39057239057239057, 4.202085937499999)(-0.3851851851851852, 3.3064453124999993)(-0.3797979797979798, 2.6115117187499997)(-0.3744107744107744, 1.8040429687499997)(-0.36902356902356903, 1.2750117187499999)(-0.36363636363636365, 0.7366992187499999)(-0.35824915824915826, 0.4791445312499999)(-0.3528619528619529, 0.24479296874999995)(-0.3474747474747475, 0.09513281249999998)(-0.3420875420875421, 0.06496874999999999)(-0.3367003367003367, 0.026683593749999995)(-0.33131313131313134, 0.010441406249999998)(-0.32592592592592595, 0.003480468749999999)(-0.32053872053872057, 0.003480468749999999)(-0.3151515151515152, 0.0011601562499999997)(-0.3097643097643098, 0.0)(-0.3043771043771044, 0.0)(-0.29898989898989903, 0.0)(-0.29360269360269364, 0.0)(-0.28821548821548826, 0.0)(-0.2828282828282829, 0.0)(-0.2774410774410775, 0.0)(-0.2720538720538721, 0.0)(-0.2666666666666667, 0.0)(-0.26127946127946133, 0.0)(-0.25589225589225595, 0.0)(-0.25050505050505056, 0.0)(-0.24511784511784518, 0.0)(-0.2397306397306398, 0.0)(-0.2343434343434344, 0.0)(-0.22895622895622902, 0.0)(-0.22356902356902353, 0.0)(-0.21818181818181814, 0.0)(-0.21279461279461276, 0.0)(-0.20740740740740737, 0.0)(-0.202020202020202, 0.0)(-0.1966329966329966, 0.0)(-0.19124579124579122, 0.0)(-0.18585858585858583, 0.0)(-0.18047138047138045, 0.0)(-0.17508417508417506, 0.0)(-0.16969696969696968, 0.0)(-0.1643097643097643, 0.0)(-0.1589225589225589, 0.0)(-0.15353535353535352, 0.0)(-0.14814814814814814, 0.0)(-0.14276094276094276, 0.0)(-0.13737373737373737, 0.0)(-0.13198653198653199, 0.0)(-0.1265993265993266, 0.0)(-0.12121212121212122, 0.0)(-0.11582491582491583, 0.0)(-0.11043771043771045, 0.0)(-0.10505050505050506, 0.0)(-0.09966329966329968, 0.0)(-0.09427609427609429, 0.0)(-0.0888888888888889, 0.0)(-0.08350168350168352, 0.0)(-0.07811447811447814, 0.0)(-0.07272727272727275, 0.0)(-0.06734006734006737, 0.0)(-0.06195286195286198, 0.0)(-0.0565656565656566, 0.0)(-0.05117845117845121, 0.0)(-0.04579124579124583, 0.0)(-0.04040404040404044, 0.0)(-0.03501683501683506, 0.0)(-0.029629629629629672, 0.0)(-0.024242424242424288, 0.0)(-0.018855218855218903, 0.0)(-0.013468013468013518, 0.0)(-0.008080808080808133, 0.0)(-0.002693602693602748, 0.0)(0.002693602693602637, 0.0)(0.008080808080808133, 0.0)(0.013468013468013518, 0.0)(0.018855218855218903, 0.0)(0.024242424242424288, 0.0)(0.029629629629629672, 0.0)(0.03501683501683506, 0.0)(0.04040404040404044, 0.0)(0.04579124579124583, 0.0)(0.05117845117845121, 0.0)(0.0565656565656566, 0.0)(0.06195286195286198, 0.0)(0.06734006734006737, 0.0)(0.07272727272727275, 0.0)(0.07811447811447814, 0.0)(0.08350168350168352, 0.0)(0.0888888888888889, 0.0)(0.09427609427609429, 0.0)(0.09966329966329968, 0.0)(0.10505050505050506, 0.0)(0.11043771043771045, 0.0)(0.11582491582491583, 0.0)(0.12121212121212122, 0.0)(0.1265993265993266, 0.0)(0.13198653198653199, 0.0)(0.13737373737373737, 0.0)(0.14276094276094276, 0.0)(0.14814814814814814, 0.0)(0.15353535353535352, 0.0)(0.1589225589225589, 0.0)(0.1643097643097643, 0.0)(0.16969696969696968, 0.0)(0.17508417508417506, 0.0)(0.18047138047138045, 0.0)(0.18585858585858583, 0.0)(0.19124579124579122, 0.0)(0.1966329966329966, 0.0)(0.202020202020202, 0.0)(0.20740740740740748, 0.0)(0.21279461279461276, 0.0)(0.21818181818181825, 0.0)(0.22356902356902353, 0.0)(0.22895622895622902, 0.0)(0.2343434343434343, 0.0)(0.2397306397306398, 0.0)(0.24511784511784507, 0.0)(0.25050505050505056, 0.0)(0.25589225589225584, 0.0)(0.26127946127946133, 0.0)(0.2666666666666666, 0.0)(0.2720538720538721, 0.0)(0.2774410774410774, 0.0)(0.2828282828282829, 0.0)(0.28821548821548815, 0.0)(0.29360269360269364, 0.0)(0.2989898989898989, 0.0)(0.3043771043771044, 0.0)(0.3097643097643097, 0.0)(0.3151515151515152, 0.0)(0.32053872053872046, 0.0011601562499999997)(0.32592592592592595, 0.008121093749999999)(0.3313131313131312, 0.029003906249999996)(0.3367003367003367, 0.038285156249999994)(0.342087542087542, 0.12993749999999998)(0.3474747474747475, 0.22391015624999996)(0.352861952861953, 0.4211367187499999)(0.35824915824915826, 0.7645429687499998)(0.36363636363636376, 1.2100429687499998)(0.36902356902356903, 1.8875742187499995)(0.3744107744107745, 2.7530507812499994)(0.3797979797979798, 3.372574218749999)(0.3851851851851853, 4.182363281249999)(0.39057239057239057, 4.6243828124999995)(0.39595959595959607, 5.044359374999999)(0.40134680134680134, 4.803046874999999)(0.40673400673400684, 4.513007812499999)(0.4121212121212121, 3.842437499999999)(0.4175084175084176, 2.9943632812499996)(0.4228956228956229, 2.2228593749999996)(0.4282828282828284, 1.5105234374999996)(0.43367003367003365, 0.9652499999999998)(0.43905723905723915, 0.5905195312499999)(0.4444444444444444, 0.31904296874999993)(0.4498316498316499, 0.16706249999999997)(0.4552188552188552, 0.08701171874999998)(0.4606060606060607, 0.025523437499999996)(0.46599326599326596, 0.026683593749999995)(0.47138047138047146, 0.003480468749999999)(0.47676767676767673, 0.0011601562499999997)(0.4821548821548822, 0.0)(0.4875420875420875, 0.0)(0.492929292929293, 0.0)(0.49831649831649827, 0.0)(0.5037037037037038, 0.0011601562499999997)(0.509090909090909, 0.0)(0.5144781144781145, 0.0011601562499999997)(0.5198653198653198, 0.004640624999999999)(0.5252525252525253, 0.005800781249999999)(0.5306397306397306, 0.019722656249999995)(0.5360269360269361, 0.05568749999999999)(0.5414141414141413, 0.11485546874999998)(0.5468013468013468, 0.21694921874999995)(0.5521885521885521, 0.3921328124999999)(0.5575757575757576, 0.7250976562499999)(0.5629629629629629, 1.0998281249999997)(0.5683501683501684, 1.7425546874999998)(0.5737373737373737, 2.4896953124999994)(0.5791245791245792, 3.4166601562499994)(0.5845117845117846, 3.9688945312499992)(0.5898989898989899, 4.559414062499999)(0.5952861952861954, 4.876136718749999)(0.6006734006734007, 4.902820312499999)(0.6060606060606062, 4.515328124999999)(0.6114478114478115, 3.852878906249999)(0.616835016835017, 3.1219804687499995)(0.6222222222222222, 2.3377148437499997)(0.6276094276094277, 1.5348867187499997)(0.632996632996633, 1.0139765625)(0.6383838383838385, 0.5731171874999998)(0.6437710437710438, 0.30628124999999995)(0.6491582491582493, 0.17054296874999997)(0.6545454545454545, 0.09165234374999998)(0.65993265993266, 0.029003906249999996)(0.6653198653198653, 0.012761718749999998)(0.6707070707070708, 0.003480468749999999)(0.6760942760942761, 0.0011601562499999997)(0.6814814814814816, 0.0)(0.6868686868686869, 0.0)(0.6922558922558923, 0.0)(0.6976430976430976, 0.0)(0.7030303030303031, 0.0)(0.7084175084175084, 0.0)(0.7138047138047139, 0.0)(0.7191919191919192, 0.0)(0.7245791245791247, 0.0)(0.7299663299663299, 0.0)(0.7353535353535354, 0.0)(0.7407407407407407, 0.0)(0.7461279461279462, 0.0)(0.7515151515151515, 0.0)(0.756902356902357, 0.0)(0.7622895622895622, 0.0)(0.7676767676767677, 0.0)(0.773063973063973, 0.0)(0.7784511784511785, 0.0)(0.7838383838383838, 0.0)(0.7892255892255893, 0.0)(0.7946127946127945, 0.0)
            };
        \legend{2-code det. HQA, True density}
        \end{axis}
\end{tikzpicture}

%% file: tikzdiagrams/toy_data_graph_3.tex
\begin{tikzpicture}[xscale=0.46,yscale=0.46,every node/.append style={font=\Large}]
        \begin{axis}[
            ylabel near ticks,
            xlabel={\huge$x$},
            xmin=-0.8, xmax=0.8,
            ymin=-0.2, ymax=11,
            xticklabel style={/pgf/number format/fixed},
            ticks=none,
            legend pos=north east,
            ymajorgrids=false,
        ]
        \addplot[,color=mycolour2,mark size=4pt,]
            coordinates {
                (-0.7973063973063974, 0.0)(-0.7919191919191919, 0.0)(-0.7865319865319866, 0.0)(-0.7811447811447811, 0.0)(-0.7757575757575759, 0.0)(-0.7703703703703704, 0.0)(-0.7649831649831651, 0.0)(-0.7595959595959596, 0.0)(-0.7542087542087543, 0.0)(-0.7488215488215488, 0.0)(-0.7434343434343436, 0.0)(-0.7380471380471381, 0.0)(-0.7326599326599328, 0.0)(-0.7272727272727273, 0.0)(-0.721885521885522, 0.0)(-0.7164983164983165, 0.0)(-0.7111111111111112, 0.0)(-0.7057239057239058, 0.0)(-0.7003367003367004, 0.0)(-0.694949494949495, 0.0)(-0.6895622895622896, 0.0)(-0.6841750841750842, 0.0)(-0.6787878787878788, 0.0)(-0.6734006734006734, 0.0)(-0.6680134680134681, 0.0)(-0.6626262626262627, 0.0)(-0.6572390572390573, 0.0)(-0.6518518518518519, 0.0)(-0.6464646464646464, 0.0)(-0.6410774410774411, 0.0)(-0.6356902356902356, 0.0)(-0.6303030303030304, 0.0)(-0.6249158249158249, 0.0)(-0.6195286195286196, 0.0)(-0.6141414141414141, 0.0)(-0.6087542087542088, 0.0)(-0.6033670033670033, 0.0)(-0.5979797979797981, 0.0)(-0.5925925925925926, 0.0)(-0.5872053872053873, 0.0)(-0.5818181818181818, 0.0)(-0.5764309764309765, 0.0)(-0.571043771043771, 0.0)(-0.5656565656565657, 0.0)(-0.5602693602693603, 0.0)(-0.554882154882155, 0.0)(-0.5494949494949495, 0.0)(-0.5441077441077442, 0.017753125)(-0.5387205387205387, 0.014015625000000002)(-0.5333333333333334, 0.0)(-0.5279461279461279, 0.03363750000000001)(-0.5225589225589227, 0.10932187500000001)(-0.5171717171717172, 0.30834375)(-0.5117845117845118, 0.8278562500000001)(-0.5063973063973064, 1.1352656250000002)(-0.501010101010101, 1.9425656250000003)(-0.49562289562289563, 3.3824375000000004)(-0.49023569023569025, 4.384087500000001)(-0.48484848484848486, 4.977415625000001)(-0.4794612794612795, 5.481978125)(-0.4740740740740741, 5.157750000000001)(-0.4686868686868687, 4.783065625000001)(-0.4632996632996633, 3.3170312500000003)(-0.45791245791245794, 2.376115625)(-0.45252525252525255, 1.4557562500000003)(-0.44713804713804717, 0.7288125000000001)(-0.4417508417508418, 0.5951968750000001)(-0.4363636363636364, 0.28685312500000004)(-0.430976430976431, 0.0934375)(-0.42558922558922563, 0.013081250000000003)(-0.42020202020202024, 0.0)(-0.41481481481481486, 0.0)(-0.4094276094276095, 0.0)(-0.4040404040404041, 0.0)(-0.3986531986531987, 0.0)(-0.39326599326599326, 0.0)(-0.3878787878787879, 0.0)(-0.3824915824915825, 0.0)(-0.3771043771043771, 0.010278125)(-0.3717171717171717, 0.010278125)(-0.36632996632996634, 0.029900000000000003)(-0.36094276094276095, 0.06073437500000001)(-0.35555555555555557, 0.15323750000000003)(-0.3501683501683502, 0.44009062500000007)(-0.3447811447811448, 0.9624062500000001)(-0.3393939393939394, 1.337090625)(-0.33400673400673403, 2.084590625)(-0.32861952861952864, 3.6309812500000005)(-0.32323232323232326, 3.7309593750000003)(-0.3178451178451179, 4.63263125)(-0.3124579124579125, 4.061728125)(-0.3070707070707071, 3.9187687500000004)(-0.3016835016835017, 3.1890218750000003)(-0.29629629629629634, 1.9332218750000003)(-0.29090909090909095, 1.4277250000000001)(-0.28552188552188557, 0.8269218750000001)(-0.2801346801346802, 0.46625312500000005)(-0.2747474747474748, 0.264428125)(-0.2693602693602694, 0.09810937500000001)(-0.263973063973064, 0.0710125)(-0.25858585858585864, 0.0)(-0.25319865319865326, 0.0)(-0.24781144781144787, 0.0)(-0.2424242424242425, 0.0)(-0.2370370370370371, 0.0)(-0.23164983164983172, 0.0)(-0.22626262626262628, 0.0)(-0.22087542087542084, 0.0)(-0.21548821548821545, 0.0)(-0.21010101010101007, 0.0)(-0.20471380471380468, 0.0)(-0.1993265993265993, 0.0)(-0.1939393939393939, 0.0)(-0.18855218855218853, 0.0)(-0.18316498316498314, 0.0)(-0.17777777777777776, 0.0)(-0.17239057239057237, 0.0)(-0.167003367003367, 0.0)(-0.1616161616161616, 0.0)(-0.15622895622895622, 0.0)(-0.15084175084175083, 0.0)(-0.14545454545454545, 0.0)(-0.14006734006734006, 0.0)(-0.13468013468013468, 0.0)(-0.1292929292929293, 0.0)(-0.12390572390572391, 0.0)(-0.11851851851851852, 0.0)(-0.11313131313131314, 0.0)(-0.10774410774410775, 0.0)(-0.10235690235690237, 0.0)(-0.09696969696969698, 0.0)(-0.0915824915824916, 0.0)(-0.08619528619528621, 0.0)(-0.08080808080808083, 0.0)(-0.07542087542087544, 0.0)(-0.07003367003367006, 0.0)(-0.06464646464646467, 0.0)(-0.05925925925925929, 0.0)(-0.053872053872053904, 0.0)(-0.04848484848484852, 0.0)(-0.043097643097643135, 0.0)(-0.03771043771043775, 0.0)(-0.032323232323232365, 0.0)(-0.02693602693602698, 0.0)(-0.021548821548821595, 0.0)(-0.01616161616161621, 0.0)(-0.010774410774410825, 0.0)(-0.00538720538720544, 0.0)(-5.551115123125783e-17, 0.0)(0.005387205387205385, 0.0)(0.010774410774410825, 0.0)(0.01616161616161621, 0.0)(0.021548821548821595, 0.0)(0.02693602693602698, 0.0)(0.032323232323232365, 0.0)(0.03771043771043775, 0.0)(0.043097643097643135, 0.0)(0.04848484848484852, 0.0)(0.053872053872053904, 0.0)(0.05925925925925929, 0.0)(0.06464646464646467, 0.0)(0.07003367003367006, 0.0)(0.07542087542087544, 0.0)(0.08080808080808083, 0.0)(0.08619528619528621, 0.0)(0.0915824915824916, 0.0)(0.09696969696969698, 0.0)(0.10235690235690237, 0.0)(0.10774410774410775, 0.0)(0.11313131313131314, 0.0)(0.11851851851851852, 0.0)(0.12390572390572391, 0.0)(0.1292929292929293, 0.0)(0.13468013468013468, 0.0)(0.14006734006734006, 0.0)(0.14545454545454545, 0.0)(0.15084175084175083, 0.0)(0.15622895622895622, 0.0)(0.1616161616161616, 0.0)(0.167003367003367, 0.0)(0.17239057239057237, 0.0)(0.17777777777777776, 0.0)(0.18316498316498314, 0.0)(0.18855218855218853, 0.0)(0.1939393939393939, 0.0)(0.1993265993265993, 0.0)(0.20471380471380474, 0.0)(0.21010101010101012, 0.0)(0.2154882154882155, 0.0)(0.2208754208754209, 0.0)(0.22626262626262628, 0.0)(0.23164983164983166, 0.0)(0.23703703703703705, 0.0)(0.24242424242424243, 0.0)(0.24781144781144782, 0.0)(0.2531986531986532, 0.0)(0.2585858585858586, 0.0)(0.26397306397306397, 0.0)(0.26936026936026936, 0.014015625000000002)(0.27474747474747474, 0.015884375000000003)(0.2801346801346801, 0.024293750000000003)(0.2855218855218855, 0.05232500000000001)(0.2909090909090909, 0.13641875)(0.2962962962962963, 0.4513031250000001)(0.30168350168350166, 0.9596031250000001)(0.30707070707070705, 1.3651218750000003)(0.31245791245791243, 2.1677500000000003)(0.3178451178451178, 3.4338281250000007)(0.3232323232323232, 4.002862500000001)(0.3286195286195286, 4.602731250000001)(0.334006734006734, 4.413053125)(0.33939393939393936, 4.328959375)(0.34478114478114474, 3.3422593750000003)(0.35016835016835024, 2.1191625000000003)(0.3555555555555556, 1.4959343750000003)(0.360942760942761, 0.8923281250000001)(0.3663299663299664, 0.5083000000000001)(0.3717171717171718, 0.22985625000000004)(0.37710437710437716, 0.08315937500000001)(0.38249158249158255, 0.06353750000000001)(0.38787878787878793, 0.0)(0.3932659932659933, 0.0)(0.3986531986531987, 0.0)(0.4040404040404041, 0.0)(0.4094276094276095, 0.0)(0.41481481481481486, 0.0)(0.42020202020202024, 0.0)(0.42558922558922563, 0.017753125)(0.430976430976431, 0.019621875000000004)(0.4363636363636364, 0.031768750000000005)(0.4417508417508418, 0.05512812500000001)(0.44713804713804717, 0.09624062500000001)(0.45252525252525255, 0.5036281250000001)(0.45791245791245794, 1.01846875)(0.4632996632996633, 1.3184031250000001)(0.4686868686868687, 2.4405875000000004)(0.4740740740740741, 3.9720281250000005)(0.4794612794612795, 4.4457562500000005)(0.48484848484848486, 5.055903125)(0.49023569023569025, 5.468896875)(0.49562289562289563, 4.744756250000001)(0.501010101010101, 4.196278125000001)(0.5063973063973064, 2.4630125000000005)(0.5117845117845118, 1.9369593750000003)(0.5171717171717172, 1.0278125000000002)(0.5225589225589226, 0.63350625)(0.5279461279461279, 0.3989781250000001)(0.5333333333333333, 0.09530625000000001)(0.5387205387205387, 0.09437187500000001)(0.5441077441077441, 0.0)(0.5494949494949495, 0.0)(0.5548821548821549, 0.0)(0.5602693602693603, 0.0)(0.5656565656565656, 0.0)(0.571043771043771, 0.0)(0.5764309764309764, 0.0)(0.5818181818181819, 0.0)(0.5872053872053873, 0.0)(0.5925925925925927, 0.0)(0.5979797979797981, 0.0)(0.6033670033670034, 0.0)(0.6087542087542088, 0.0)(0.6141414141414142, 0.0)(0.6195286195286196, 0.0)(0.624915824915825, 0.0)(0.6303030303030304, 0.0)(0.6356902356902357, 0.0)(0.6410774410774411, 0.0)(0.6464646464646465, 0.0)(0.6518518518518519, 0.0)(0.6572390572390573, 0.0)(0.6626262626262627, 0.0)(0.6680134680134681, 0.0)(0.6734006734006734, 0.0)(0.6787878787878788, 0.0)(0.6841750841750842, 0.0)(0.6895622895622896, 0.0)(0.694949494949495, 0.0)(0.7003367003367004, 0.0)(0.7057239057239058, 0.0)(0.7111111111111111, 0.0)(0.7164983164983165, 0.0)(0.7218855218855219, 0.0)(0.7272727272727273, 0.0)(0.7326599326599327, 0.0)(0.7380471380471381, 0.0)(0.7434343434343434, 0.0)(0.7488215488215488, 0.0)(0.7542087542087542, 0.0)(0.7595959595959596, 0.0)(0.764983164983165, 0.0)(0.7703703703703704, 0.0)(0.7757575757575758, 0.0)(0.7811447811447811, 0.0)(0.7865319865319865, 0.0)(0.7919191919191919, 0.0)(0.7973063973063973, 0.0)
            };
        
        \legend{2-code stoch. HQA}
        \end{axis}
\end{tikzpicture}

%% file: tikzdiagrams/method-diagram.tex
\begin{tikzpicture}
    \node [draw,rectangle, inner sep=0.01cm] (input) {\includegraphics[width=0.3\linewidth]{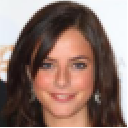}};
    \node [draw, thick, color=OliveGreen!50,trapezium,fill=OliveGreen!20, text=OliveGreen, minimum height=1.2cm, minimum width=2.5cm] (encoder0) [ above=of input] {\large encoder};
    \node [draw,color=Blue!50,trapezium,fill=CornflowerBlue!20, text=Blue, minimum height=1.2cm, minimum width=2.5cm] (decoder0) [right=of encoder0, xshift=1cm] {\large decoder};
    \node [draw, rectangle, inner sep=0.01cm] (output0) [below=of decoder0]{\includegraphics[width=0.3\linewidth]{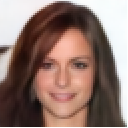}};
    \node [inner sep=0] (quantize0) [above=of decoder0] {\includegraphics[scale=0.3]{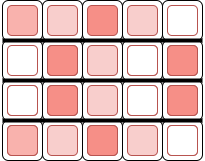}};

    \coordinate [above=of encoder0, yshift=1cm] (topleft0) {};
    \coordinate[above=of topleft0, xshift=-2.5cm] (parting0left) {};
    \coordinate[right=of parting0left, xshift=10cm] (parting0right) {};
    
    \node[] (layer0text) [below right=of parting0left, xshift=-1cm, yshift=1cm] {\large Layer 1};
    \node [draw, thick, color=OliveGreen!50,trapezium,fill=OliveGreen!20, text=OliveGreen, minimum width = 3cm, minimum height=.75cm] (encoder1) [ above=of topleft0, yshift=3cm] {\large encoder};
    \node[inner sep=0]  (embeddinginput1) [below=of encoder1] {\includegraphics[scale=0.3]{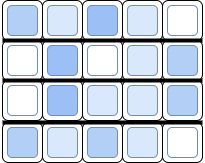}};
    \node[inner sep=0]  (embedding1) [above right=of embeddinginput1, yshift=1.4cm] {\includegraphics[scale=0.3]{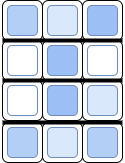}};
    \node [inner sep=0] (quantize1) [right=of embedding1, xshift=1.15cm] {\includegraphics[scale=0.3]{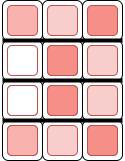}};
    \node [draw,color=Blue!50,trapezium,fill=CornflowerBlue!20, text=Blue, minimum height=.75cm, minimum width=3cm] (decoder1) [right=of encoder1, xshift=2.55cm] {\large decoder};
    \node[inner sep=0]  (embeddingoutput1) [right=of embeddinginput1, xshift=3.11cm] {\includegraphics[scale=0.3]{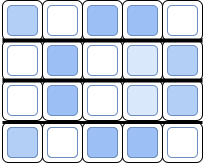}};
    \coordinate [above=of encoder1, yshift=0.47cm] (topleft1) {};
    \coordinate[above=of topleft1, xshift=-2.5cm] (parting1left) {};
    \coordinate[right=of parting1left, xshift=10cm] (parting1right) {};
    \coordinate[above=of embedding1, yshift=-.5cm] (MSE2) {};
    \coordinate[above=of MSE2] (topdiagram) {};
    
    \node[] (layer1text) [below right=of parting1left, xshift=-1cm, yshift=1cm] {\large Layer 2};
    
    \draw [-] (input.north west) -- (encoder0.bottom left corner);
    \draw [-] (input.north east) -- (encoder0.bottom right corner);
    
    \draw [->] (encoder0) -- (embeddinginput1);
    \draw [-] (output0.north west) -- (decoder0.bottom left corner);
    \draw [-] (output0.north east) -- (decoder0.bottom right corner);
    
    \draw [transform canvas={xshift=0.25cm}, ->, red] (embeddingoutput1) -- (quantize0);
    \draw [transform canvas={xshift=0.65cm}, ->, red] (embeddingoutput1) -- (quantize0);
    \draw [transform canvas={xshift=-0.25cm}, ->, red] (embeddingoutput1) -- (quantize0);
    \draw [transform canvas={xshift=-0.65cm}, ->, red] (embeddingoutput1) -- (quantize0);
    
    \draw [->] (quantize0) -- (decoder0);

    \draw [<->, dashed] (embeddinginput1) -- (embeddingoutput1) node [midway, above] (MSE) {\large MSE};

    \draw [-] (embeddinginput1.north west) -- (encoder1.bottom left corner);
    \draw [-] (embeddinginput1.north east) -- (encoder1.bottom right corner);
    \draw [-] (embeddingoutput1.north west) -- (decoder1.bottom left corner);
    \draw [-] (embeddingoutput1.north east) -- (decoder1.bottom right corner);
    
    \draw [-] (encoder1) -- (topleft1);
    \draw [->] (topleft1) -- (embedding1);
    
    \draw [transform canvas={yshift=0.25cm}, ->, red] (embedding1) -- (quantize1);
    \draw [transform canvas={yshift=0.65cm}, ->, red] (embedding1) -- (quantize1);
    \draw [transform canvas={yshift=-0.25cm}, ->, red] (embedding1) -- (quantize1);
    \draw [transform canvas={yshift=-0.65cm}, ->, red] (embedding1) -- (quantize1);
    
    \draw [->] (quantize1) -- (decoder1);
    
\end{tikzpicture}

%% file: tikzdiagrams/ShallowMoG.tex
\begin{tikzpicture}

  \node[obs]                               (x) {$x$};
  \node[latent, above=of x] (z) {$z$};
  \node[const, above right=of x, red, yshift=-0.3cm, xshift=-0.3cm] (qz) {$q(z|x)$};
  \node[const, above left=of x, yshift=-0.3cm, xshift=0.8cm] (pze) {$p(x|z)$};
  \edge {z} {x} ; 
  
  \draw[->, dashed]
    (x) [out=45, in=315,red] to  (z);

\end{tikzpicture}

%% file: tikzdiagrams/DeepMoG.tex
\begin{tikzpicture}

  \node[obs]                               (x) {$x$};
  \node[latent, above=of x] (ze) {$z_e$};
  \node[latent, above=of ze] (z) {$z$};
  \node[const, above right=of x, red, yshift=-0.3cm,xshift=-0.3cm] (qze) {$q(z_e|x)$};
  \node[const, above right=of ze, red, yshift=-0.3cm, xshift=-0.3cm] (qz) {$q(z|z_e)$};
  \node[const, above left=of ze, yshift=-0.3cm, xshift=0.8cm] (pze) {$p(z_e|z)$};
  \node[const, above left=of x, yshift=-0.3cm, xshift=0.8cm] (px) {$p(x|z_e)$};
  \edge {ze} {x} ; %
  \edge {z} {ze};
  
  \draw[->, dashed]
    (x) [out=45, in=315,red] to  (ze);
  \draw[->, dashed]    
    (ze) [out=45, in=315,red] to  (z);

\end{tikzpicture}